\documentclass[sigconf]{acmart}
\AtBeginDocument{%
  }

\setcopyright{acmlicensed}
\copyrightyear{2026}
\acmYear{2026}
\acmDOI{XXXXXXX.XXXXXXX}
\acmConference[Conference acronym ’XX]{Make sure to enter the correct
  conference title from your rights confirmation email}{June 03--05,
  2018}{Woodstock, NY}
\acmISBN{978-1-4503-XXXX-X/2018/06}

\usepackage{booktabs}
\usepackage{subcaption}
\usepackage{tcolorbox}
\usepackage{multirow}
\usepackage{array}
\usepackage{mdframed}
\usepackage{xcolor}
\usepackage{placeins}
\usepackage{soul} 
\usepackage{xcolor}
\usepackage{courier} 

\definecolor{pastelblue}{HTML}{7BAAF7}
\definecolor{pastelgreen}{HTML}{A8E6CF}
\definecolor{pastelpurple}{HTML}{C292E2}
\definecolor{pastelred}{HTML}{FF8A80}

\definecolor{darkblue}{HTML}{1565C0}
\definecolor{darkgreen}{HTML}{388E3C}
\definecolor{darkpurple}{HTML}{6A1B9A}
\definecolor{darkred}{HTML}{C62828}

\begin{document}

\title{A Modular Multi-task Reasoning Framework Integrating Spatio-temporal Models and LLMs}


\author{Kethmi Hirushini Hettige}
\email{kethmihi001@e.ntu.edu.sg}
\affiliation{%
  \institution{Nanyang Technological University}
  \country{Singapore}
}
\affiliation{%
  \institution{A*STAR Institute of Advanced
Intelligence and Computing}
  \country{Singapore}
  }

\author{Jiahao Ji}
\email{jiahaoji03@gmail.com}
\affiliation{%
  \institution{Beihang University}
  \country{China}
  }
\affiliation{%
  \institution{Meituan Inc.}
  \city{Beijing}
  \country{China}
  }

\author{Cheng Long}
\email{c.long@ntu.edu.sg}
\affiliation{%
  \institution{Nanyang Technological University}
  \country{Singapore}
  }

\author{Shili Xiang}
\email{Xiang\_Shili@a-star.edu.sg}
\affiliation{%
  \institution{A*STAR Institute of Advanced
Intelligence and Computing}
  \country{Singapore}
  }

\author{Gao Cong}
\email{gaocong@ntu.edu.sg}
\affiliation{%
  \institution{Nanyang Technological University}
  \country{Singapore}
  }

\author{Jingyuan Wang}
\email{jywang@buaa.edu.cn}
\affiliation{%
  \institution{Beihang University}
  \country{China}
  }

\renewcommand{\shortauthors}{Hettige et al.}

\begin{abstract}
Spatio-temporal data mining plays a pivotal role in informed decision making across diverse domains. However, existing models are often restricted to narrow tasks, lacking the capacity for multi-task inference and complex long-form reasoning that requires generation of in-depth, explanatory outputs. These limitations restrict their applicability to real-world, multi-faceted decision scenarios. In this work, we introduce STReason, a novel framework that integrates the reasoning strengths of large language models (LLMs) with the analytical capabilities of spatio-temporal models for multi-task inference and execution. Without task-specific fine-tuning, STReason leverages in-context learning to decompose complex natural language queries into modular, interpretable programs, which are then systematically executed to generate both numerical solutions and detailed reasoning rationales. By grounding all explanations in verified computational outputs, STReason inherently suppresses factual hallucinations common in direct-prompt LLM approaches. To enable rigorous evaluation, we construct a new benchmark dataset and propose a unified evaluation framework with metrics specifically designed for long-form spatio-temporal reasoning. Experimental results demonstrate that STReason significantly outperforms advanced LLM baselines across all reasoning metrics, particularly excelling in complex, reasoning-intensive spatio-temporal scenarios. Human evaluations further validate STReason’s credibility and practical utility, demonstrating its potential to reduce expert workload and broaden the applicability to real-world spatio-temporal tasks. Our code is available at: \url{https://github.com/kethmih/STReason}.
\end{abstract}





\maketitle

\section{Introduction} \label{sec:Introduction}

In the realm of data science, spatio-temporal data, characterized by both spatial and temporal dimensions, plays a critical role in a wide array of fields such as environmental monitoring \citep{hettige2024airphynet,liang2023airformer}, urban planning and traffic management \citep{li2024urbangpt,ji2022stden}, and public health \citep{dong2024brain}. Over the years, research in spatio-temporal data mining has progressed from conventional statistical and machine learning approaches \citep{xie2020urban} to advanced deep learning frameworks \citep{jin2023spatio,wang2020deep,zhang2024survey}. In recent years, the development of Foundation Models (FMs) has sparked a surge in research aimed at improving spatio-temporal modeling through Large Language Models (LLMs) \citep{li2024urbangpt,liang2024foundation,liang2025foundation,li2025stbench,zhang2024survey}. By leveraging the strengths of LLMs in generalization, cross-modal reasoning, and long-sequence modeling, several LLM-based spatio-temporal models have been developed for various applications with notable performance improvements in zero-shot and few-shot scenarios \citep{cao2023tempo,zhou2023onefitsall,chen2023gatgpt,DBLP:conf/dsaa/AlnegheimishNBV24}.

Despite significant advancements, current LLM-based spatio-temporal models exhibit critical limitations. \textbf{First}, while these models are commonly applied for numerical tasks such as forecasting, their use in inferential problem-solving, such as reasoning or decision-making, remains underexplored \citep{li2024urbangpt,zhou2023onefitsall,zhang2023promptst,yuan2024unist}. For example, a forecasting system might violate real-world constraints, and erroneously predict traffic speeds that exceed safety thresholds, thus limiting interpretability and reliability of their outputs. \textbf{Second}, existing LLM-based spatio-temporal models predominantly rely on task-specific fine-tuning or parameter-efficient updates \citep{li2024urbangpt, tan2024language, kambhampati2024position}, incurring 
substantial computational costs, risking catastrophic 
forgetting, and requiring retraining whenever the LLM 
backbone is updated. This tightly couples model performance to a fixed training distribution, limiting adaptability to new domains or query types without significant re-engineering effort. \textbf{Third}, most current models are restricted to fixed spatio-temporal input formats; typically tensors with pre-defined dimensions (e.g. [batch, time, location, feature]) and struggle with processing natural language queries, highlighting a gap in their utility as general-purpose AI systems \citep{zhou2023onefitsall,chen2023gatgpt,liu2024spatial,liu2024timecma}. \textbf{Fourth}, no standardized benchmark exists for evaluating multi-task, long-form spatio-temporal reasoning and existing evaluation protocols assess narrow, single-task outputs such as scalar forecasts or binary 
labels \citep{li2024urbangpt, yuan2024unist}, which are 
not suited for assessing comprehensive, explanatory responses.


Building on these limitations, few recent studies have explored models for spatio-temporal reasoning, though they predominantly focus on highly task-specific inference problems. The majority of these approaches convert spatio-temporal data into textual descriptions, for processing by LLMs \citep{peng2025lc, chen2024genfollower, guo2024towards}. This translation often leads to significant information loss, shallow reasoning, and inability to capture complex dependencies inherent in spatio-temporal phenomena. To tackle these limitations, certain program-based approaches have been introduced. For example, UrbanLLM \citep{jiang2024urbanllm} fine-tunes LLMs for urban planning by breaking down queries into sub-tasks handled by spatio-temporal AI models. However, its effectiveness is limited by the specific urban contexts it was trained on, which reduces generalizability, and its reliance on pre-trained models may hinder detailed geospatial understanding. Similarly, TS-Reasoner \citep{ye2024beyond} decomposes complex time-series tasks, yet it remains confined to niche time-series applications within climate and energy data, highlighting a gap in versatility and broader applicability. More generally, recent tool-augmented reasoning frameworks demonstrate the broader promise of tool-use paradigms for structured reasoning \citep{qu2025tool, fan2026tool}. However, these approaches target general-purpose or visual reasoning tasks and produce short-form outputs, without addressing the structured numerical analysis, domain-specific constraint validation, or long-form explanatory generation required for spatio-temporal decision support.  Long-form question answering \citep{fan-etal-2019-eli5} where comprehensive, explanatory, and interpretable outputs are generated from complex inputs still remains underexplored in spatio-temporal settings.
\begin{figure}[h]
    \begin{center}
    \includegraphics[width=\linewidth]{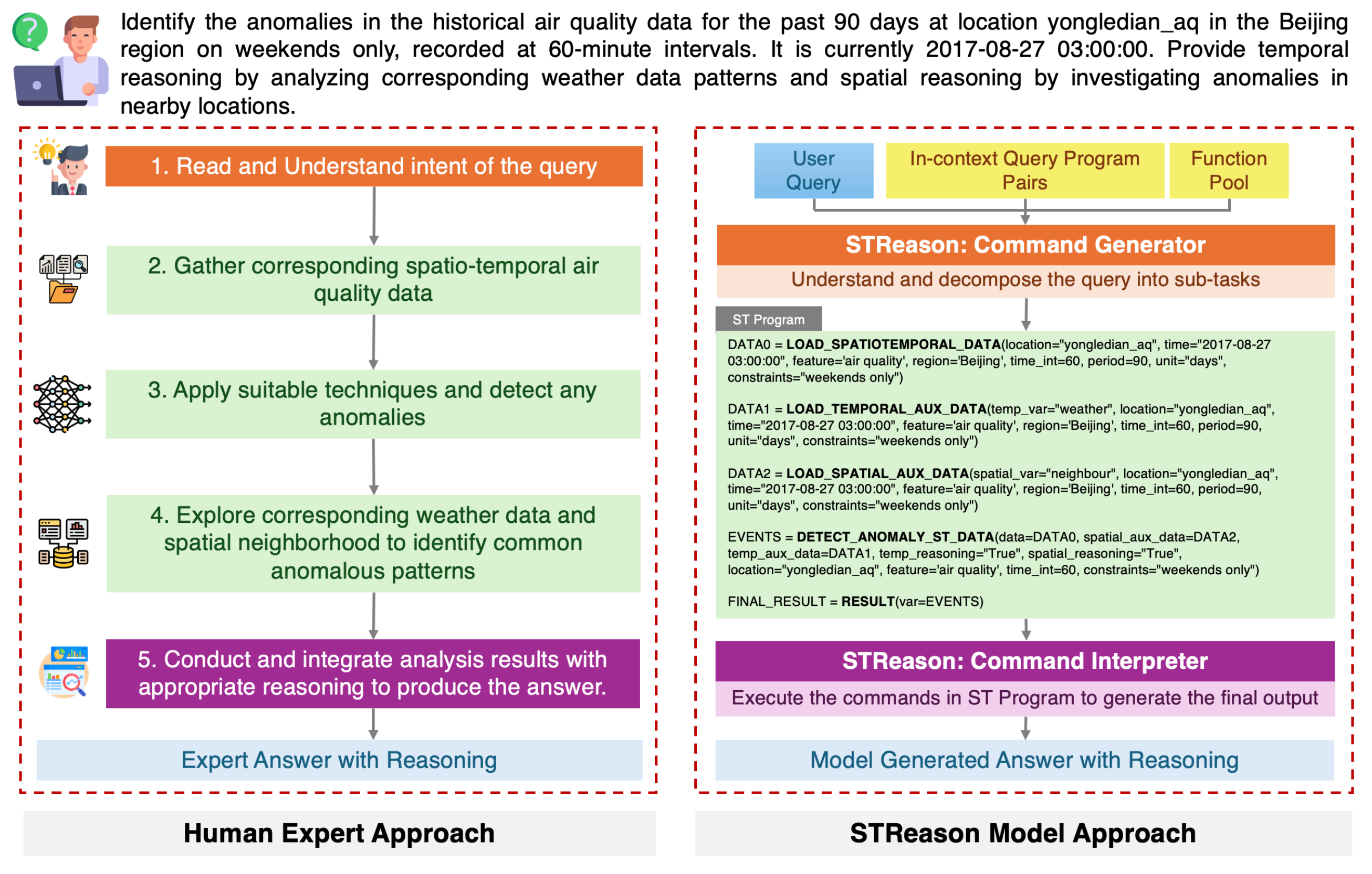}
    \end{center}
    \caption{Comparison between Human Expert and STReason Model workflows for answering a complex spatio-temporal query.}
    \label{fig:QASystem_Comparison}
\end{figure}

Addressing these limitations requires a paradigm shift to a task and domain-agnostic framework that can adapt to complex spatio-temporal problems and generate outputs with rich context, depth, and clarity making results more interpretable and actionable. Consider answering a query on analyzing historical air quality data for anomalies, integrating temporal and spatial reasoning by considering corresponding weather patterns and nearby locations as shown in Figure \ref{fig:QASystem_Comparison}. Typically, a human expert (Left)  begins by understanding the intent of the query, followed by executing a series of analytical steps and finally providing a comprehensive answer that synthesizes all gathered insights with appropriate reasoning. While existing LLM-based systems can handle these individual steps in isolation, they fall short of effectively executing complex, language driven spatio-temporal queries end-to-end. They struggle to fully comprehend inputs, integrate multi-dimensional knowledge, and produce reasoned, actionable outputs. Bridging this gap requires advanced LLM agents that function like full-stack data scientists, supporting the entire pipeline from data ingestion and analysis to interpretation and decision-making.

In response to these challenges, we introduce STReason, a novel framework that combines the reasoning and comprehension strengths of LLMs with the analytical power of state-of-the-art spatio-temporal models for multi-task inference and execution. Without requiring task-specific fine-tuning, STReason leverages LLMs to decompose complex tasks, articulated in natural language, into structured programs (``ST Program") using predefined in-context query–program example pairs and a Function Pool; a curated dictionary of available modules and their specifications that guides the LLM in aligning sub-tasks with the appropriate executable functions (see Figure \ref{fig:QASystem_Comparison} (Right)). The generated ST Programs are then executed by specialized, end-to-end trained models or tailored analytical programs to produce coherent, long-form answers that go beyond simple numeric predictions to include structured reasoning, explanatory narratives and meaningful interpretations inspired by long-form question answering \citep{fan-etal-2019-eli5}. A demonstration of STReason can be viewed at: \url{https://anon.to/T5lL94}.

STReason revolutionizes spatio-temporal data analysis by addressing the limitations of conventional LLMs or LLM agents with limited spatio-temporal understanding \citep{manvi2024geollm,shen2023hugginggpt}. Unlike task-specific models \citep{jiang2024urbanllm,li2024urbangpt}, STReason requires no pre-training, enhancing flexibility and generalization across diverse domains. Highlighting its versatility, we apply STReason to three primary tasks in the traffic and air quality domains. These tasks include (i) Spatio-temporal Analysis, (ii) Spatio-temporal Anomaly Detection, and (iii) Spatio-temporal Prediction and Reasoning. The modular nature facilitates easy customization, promoting adaptation to new tasks without significant retraining. Moreover, STReason’s structured execution allows for validation of logic and inspection of intermediate outputs crucial for high accountability domains like environmental monitoring. To enable rigorous evaluation, we also introduce a new benchmark dataset specifically designed for long form spatio-temporal reasoning. Unlike existing datasets that focus narrowly on a single task, our dataset includes multi-task natural language queries, corresponding structured programs, and a ground-truth answer annotated with the key components expected in a comprehensive response. This dataset offers a valuable foundation for systematically evaluating any spatio-temporal reasoning model. We summarize our contributions as follows:
\begin{enumerate}
    \item We propose STReason, a novel training-free framework that decomposes complex multi-faceted spatio-temporal queries into executable steps and produces interpretable and well-reasoned outputs without human intervention.
    \item We develop a benchmark dataset spanning three core tasks: Spatio-temporal Analysis, Anomaly Detection, and Prediction and Reasoning from real-world traffic and air quality data to rigorously assess any reasoning model, including STReason.
    \item We develop a systematic evaluation framework to assess long-form reasoning responses for spatio-temporal queries, based on constraint adherence, factual accuracy, and logical coherence.
    \item Extensive experiments demonstrate that STReason 
    significantly outperforms advanced LLM baselines across 
    all evaluation metrics, while delivering interpretable, 
    constraint-validated, long-form outputs that standalone 
    spatio-temporal models cannot produce.
\end{enumerate}

\section{Related Work}
\subsection{Large Language Models for Spatio-temporal Tasks}
Recent studies have explored the use of LLMs for spatio-temporal analysis, motivated by their strong representation learning and generalization capabilities \citep{jin2023large,li2025urban}. Existing approaches broadly fall into three categories; tuning-based methods, prompt-based methods, and foundation model–centric approaches. Tuning-based approaches adapt pre-trained LLMs to spatio-temporal tasks through fine-tuning or parameter-efficient updates \citep{zhou2023onefitsall,liu2024spatial,liu2024unitime}. While effective in improving task performance, these methods incur substantial training costs, risk catastrophic forgetting, and are typically restricted to fixed task formulations and data schemas. Prompt-based approaches aim to avoid retraining by reformulating spatio-temporal data as textual or structured prompts \citep{gruver2023large,zhang2023promptst,wang2023would}.However, their performance is highly sensitive to prompt design, and they struggle with multi-step analytical reasoning, constraint satisfaction, and robustness across domains. More recently, task-specific foundation models have been proposed for spatio-temporal forecasting and analytics \citep{liang2024foundation,liang2025foundation} with efforts further extending to richer urban input modalities. For instance, UrbanLLaVA \citep{feng2025urbanllava} integrates satellite imagery, trajectories, and street views through a staged training pipeline that decouples spatial reasoning from domain learning for zero-shot urban prediction. 

Although these models achieve strong predictive performance, they require extensive domain-specific training and are primarily designed for fixed-format tensor inputs and scalar outputs. Consequently, they are not well-suited for natural language interaction, multi-task inference, or long-form analytical reasoning. Overall, existing LLM-based spatio-temporal models focus predominantly on prediction accuracy under predefined input–output formats, rather than supporting interpretable, multi-step reasoning driven by natural language queries. In contrast, STReason targets spatio-temporal analysis as a reasoning and decision-support problem. Unlike tuning-based and foundation model-centric methods, it requires no task-specific training or architectural changes, accepting free-form natural language queries across diverse domains. Unlike prompt-based methods, its execution-grounded ST Programs constrain LLM generation to verified computational outputs, structurally suppressing factual hallucinations. Together, these properties position STReason as a task and domain-agnostic framework for structured program generation, analytical execution, and long-form spatio-temporal question answering.

\subsection{Reasoning with Foundation Models} 
Recent advances in foundation models (e.g., GPT-4o, DeepSeek-R1) have demonstrated substantial progress in reasoning capabilities beyond earlier versions (e.g., GPT-3), with improved context awareness and domain adaptation \citep{guo2025deepseek}. Techniques such as Chain-of-Thought prompting and its extensions encourage intermediate reasoning steps and have improved performance on logical, numerical, and compositional tasks \citep{yao2023tree,DBLP:conf/iclr/0002WSLCNCZ23}. Program-based reasoning methods further enhance reliability by decomposing natural language queries into executable programs, enabling grounded reasoning across visual, tabular, and symbolic domains \citep{gupta2023visual,wang2024chain,ye2024beyond,huang2025vision}.Specifically, VISPROG \citep{gupta2023visual} and ViperGPT 
\citep{suris2023vipergpt} target visual QA over images, while 
Chain-of-Table \citep{wang2024chain} extends program-guided 
reasoning to structured tabular data. HuggingGPT 
\citep{shen2023hugginggpt} routes general AI tasks to expert 
models through an LLM orchestrator, and Program-of-Thoughts 
\citep{chen2022program} and PAL \citep{gao2023pal} target 
mathematical and symbolic reasoning over text. In the 
spatio-temporal domain, UrbanLLM \citep{jiang2024urbanllm} 
applies program-based decomposition to urban planning tasks, 
but remains limited to specific urban contexts, produces short numeric outputs, and requires task-specific fine-tuning. While these frameworks share the principle of decomposing natural language queries into structured programs, they differ fundamentally from STReason in domain scope, input modality, output format, and execution design. Empirically, representative 
frameworks fail to process STReason's benchmark queries, returning 
null outputs or empty programs, confirming the fundamental 
domain mismatch (see Appendix~\ref{app:related_work_ext}).

The scope of reasoning has also broadened to include commonsense, numerical, and causal tasks by integrating contextual and quantitative signals \citep{li2025stbench,li2024urbangpt,zhang2023situatedgen}. Moreover, recent work has explored agent-based systems that integrate external tools and models to solve complex tasks \citep{jiawei2024large,gao2024large,lai2025ustbench}. However, they often rely on task-specific tool chains, incur non-trivial engineering overhead, and remain vulnerable to hallucinations when operating outside their training distribution \citep{qu2025tool, fan2026tool}. Fine-tuned reasoning models for time-series tasks have also been proposed \citep{kong2025time,guan2025timeomni,luo2025time,xie2024chatts,liu2025time}; however, the majority of these approaches primarily focus on univariate or multivariate temporal sequences, with limited or no explicit modeling of spatial relationships. In addition, these methods incur additional training costs and often exhibit limited generalization beyond their target datasets. STReason addresses this gap by extending program-guided reasoning to spatio-temporal domains, enabling analytical execution over structured spatio-temporal data, enforcing domain constraints, and producing interpretable long-form explanations without additional training. 

\section{Methodology} \label{sec:Methodology}
We introduce the STReason framework which operates in two primary stages: the command generation and command execution. Given a user query, in the first stage natural language queries are converted into structured commands by a Command Generator, which leverages the in-context learning abilities of LLMs. Using well-crafted query-program pairs along with a structured Function Pool of available interpreter modules, it generates a sequence of executable commands referred to as an ``ST Program" tailored to each query. Each command in the ST Program activates one of the specific modules within our framework, which include state-of-the-art spatio-temporal prediction models, language processing units and data processing sub-routines. In the second stage, the generated ST Program is executed by a Command Interpreter that maps each command to its corresponding function, generating an integrated final response (see Figure \ref{fig:QASystem_Comparison} (Right)). This modular pipeline ensures that inputs are sequentially processed and integrated to deliver comprehensive, well-reasoned outputs.

\subsection{Command Generator}
The Command Generator is responsible for translating complex natural language queries into executable ST-Programs. Specifically, it decomposes complex spatio-temporal queries into manageable sub-tasks, leveraging the  in-context learning capabilities of LLMs. To enhance grounding and reduce ambiguity in sub-task selection, STReason augments the input prompt with a Function Pool; a curated collection of function specifications that serves as a structured knowledge base for the model. 
\begin{figure}[htbp]
    \centering
    \includegraphics[width=\linewidth]{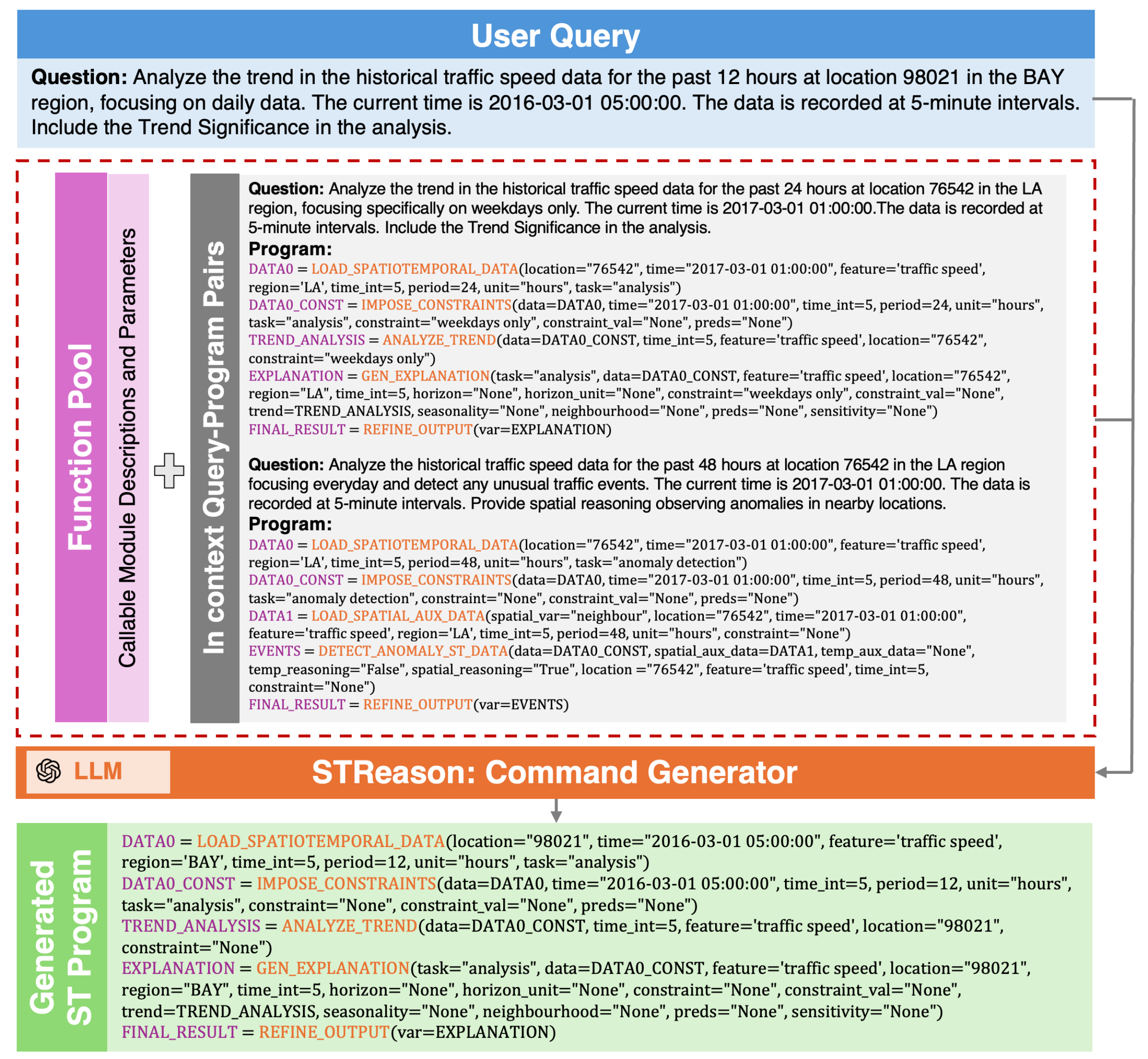}
    \caption{STReason Command Generator uses a Function Pool and in-context query-program pairs to generate executable ST-Program for a given user query.}
    \label{fig:ProgramGenerator}
\end{figure}

Each function in the pool includes its definition, input 
parameter details and output structures. Figure~\ref{fig:func_pool_structure} illustrates the structure of a representative function, which loads spatio-temporal data for a specific location and time period. Each entry specifies the call signature, a natural language description, parameter definitions, and the expected return format. This structured reference not only clarifies available modules but also helps to resolve ambiguity during program generation, particularly when in-context examples lack alignment. For example, if a user query requests seasonality analysis but the provided examples only cover trend analysis, the Function Pool guides the LLM to select the correct function (\texttt{ANALYZE\_SEASONALITY}) by referencing its defined syntax and purpose. It also improves generalization to diverse query phrasing. For example, a query like “Find anything unusual in pollution patterns over weekends” can still be correctly mapped to the anomaly detection function, despite differing wording from prior examples. 

\begin{figure}[htbp]
\centering
\begin{tcolorbox}[
    colback=gray!5!white, 
    colframe=black!70!white, 
    title=LOAD\_SPATIOTEMPORAL\_DATA, 
    fonttitle=\scriptsize \bfseries, 
    sharp corners=south,
    fontupper=\scriptsize
]
\textbf{Call Signature:} \\
\texttt{LOAD\_SPATIOTEMPORAL\_DATA(location, time, feature, region, time\_int, period, unit, task)}

\vspace{1em}
\textbf{Description:} \\
Loads spatio-temporal data for a specific location and time period.

\vspace{1em}
\textbf{Parameters:}
\begin{itemize}
    \item \texttt{location (str)} — The geographical location identifier.
    \item \texttt{time (datetime)} — The current time.
    \item \texttt{feature (str)} — Feature of interest (e.g., \texttt{'traffic speed'}, \texttt{'air quality'}).
    \item \texttt{region (str)} — The broader geographical area.
    \item \texttt{time\_int (int)} — Interval in minutes between data points.
    \item \texttt{period (int)} — Duration for which data is loaded.
    \item \texttt{unit (str)} — The time unit of the period (e.g., \texttt{'hours'}, \texttt{'days'}).
    \item \texttt{task (str)} — Task of the query (e.g., \texttt{'analysis'}, \texttt{'anomaly detection'}, \texttt{'prediction'}).
\end{itemize}

\vspace{0.5em}
\textbf{Returns:} \\
\texttt{DataFrame} containing spatio-temporal data.
\end{tcolorbox}
\caption{Structure of Sample Function of the Function Pool}
\label{fig:func_pool_structure}
\end{figure}

\begin{figure}[htbp]
    \centering
    \includegraphics[width=\linewidth]{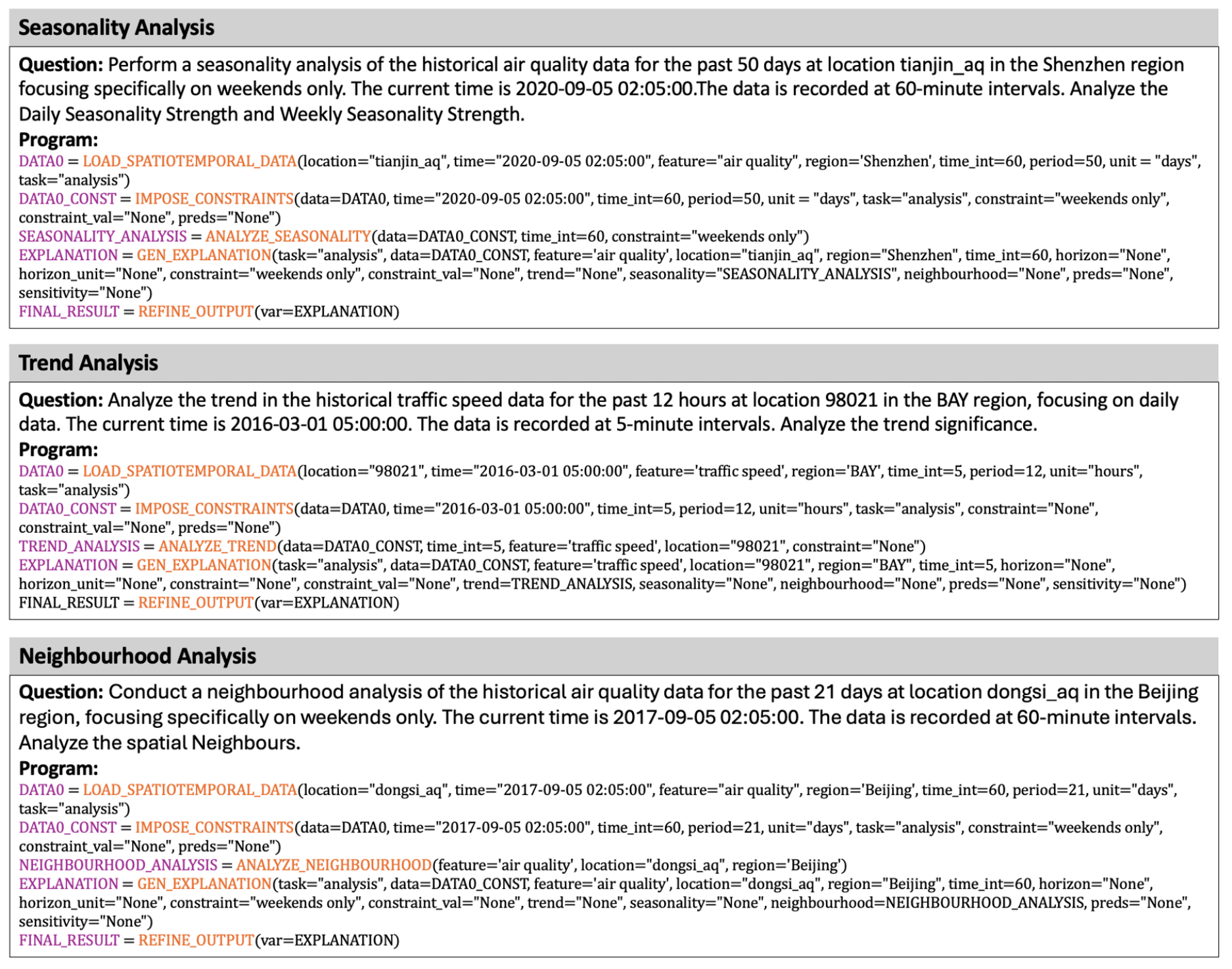}
    \caption{ST-program for Analysis Task}
    \label{fig:ProgramGen_Analysis}
\end{figure}

Together with the input query–program pairs, the Function Pool enables the Command Generator to produce a clear and interpretable program-based reasoning path, as illustrated in Figure \ref{fig:ProgramGenerator}. Critically, this design introduces an implicit hallucination suppression mechanism. Since the final \texttt{GEN\_EXPLANATION} module only receives verified numerical outputs from prior execution steps as inputs, LLM generation is structurally constrained to narrate computed results rather than fabricate them. This contrasts with direct-prompt LLMs that generate unconstrained text over raw data. We use GPT-3.5 Turbo as the backbone LLM, due to the balance of its cost-efficiency and reasoning capability. Each step of the generated program (i.e. ST-program) corresponds to a specific module within the framework, designed for analytical, predictive, or inference tasks. Particularly, each command line of an ST-program includes a module name (e.g. \texttt{ANALYZE\_TREND}, \texttt{DETECT\_ANOMALY\_ST\_DATA}), input arguments (e.g., ``data", ``location"), and an output variable (e.g., ``EVENTS", ``PREDICTION"). These naming conventions help the LLM understand and map inputs to the appropriate functions accurately. The final output from the command generator is a complete, interpretable program that defines all steps, inputs, and outputs needed to answer the query. Example ST Programs for Analysis tasks are illustrated in 
Figure~\ref{fig:ProgramGen_Analysis}, with additional examples for Anomaly Detection and Prediction and Reasoning tasks provided in Appendix~\ref{app:ProgramGen_Tasks}. 

\subsection{Command Interpreter}
The Command Interpreter sequentially executes the ST-Program, functioning like a traditional programming language interpreter. It accesses a library of pre-defined modules, invoking them as required by the program. It ensures that each command is executed in the correct order and that data flows correctly between steps, maintaining consistency and accuracy throughout the process. Finally, the results from each command are integrated to build a comprehensive response to the initial query, effectively turning raw data into insightful conclusions. 
\begin{figure}[h!]
    \centering
    \includegraphics[width=\linewidth]{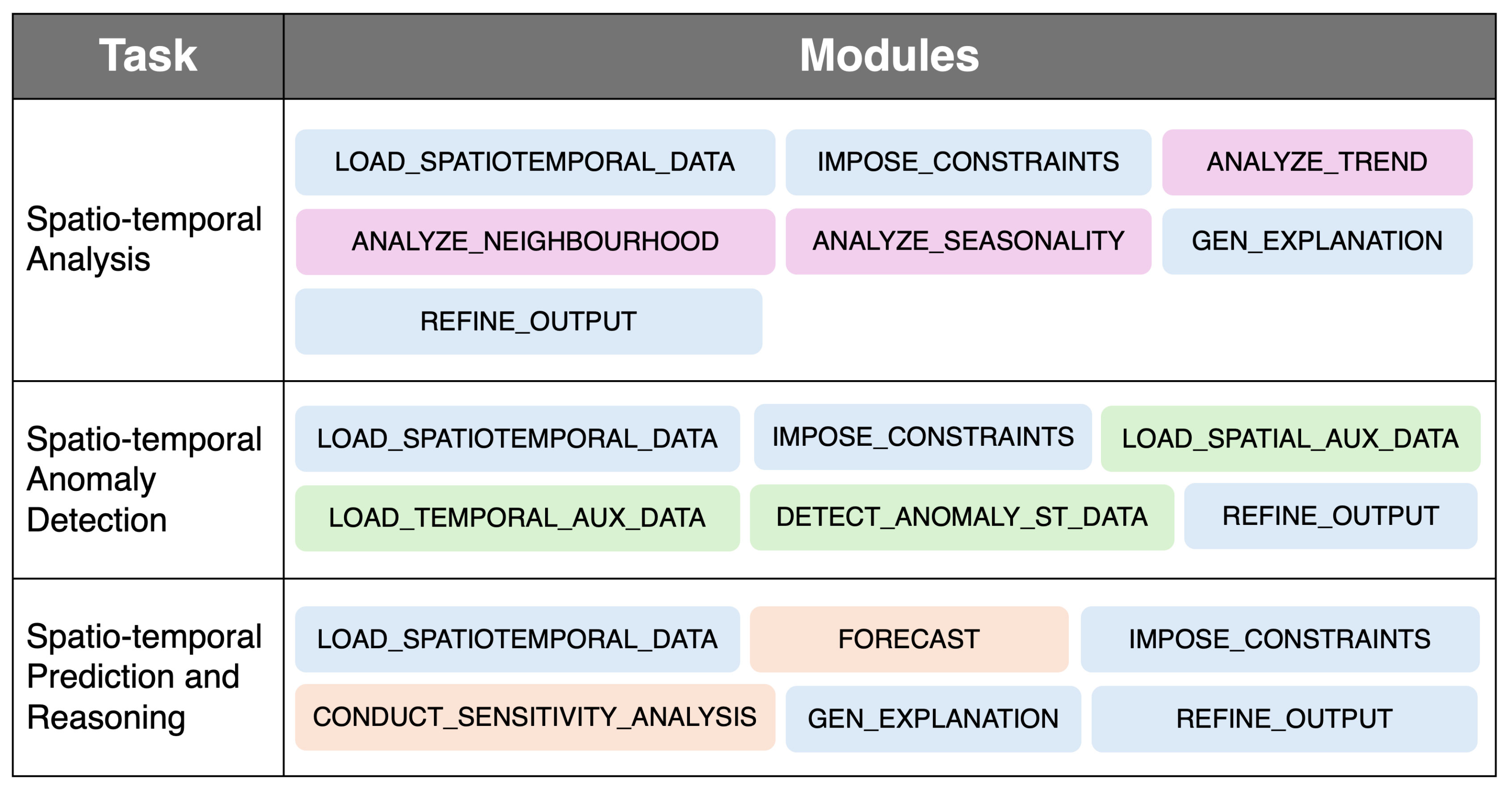}
    \caption{Command Interpreter Modules}
    \label{fig:InterpreterModules}
\end{figure}

STReason features 12 specialized modules supporting three core spatio-temporal tasks: Analysis, Anomaly Detection and Prediction and Reasoning (see Figure \ref{fig:InterpreterModules}). Inspired by VISPROG \citep{gupta2023visual}, each module is implemented as a Python class with methods for parsing inputs, executing computations, and summarizing outputs. Full module details are provided in Appendix \ref{app:extended_appendix}. This modular architecture simplifies the integration of new modules into STReason, requiring only the development and registration of a new module class. To further enhance transparency and user comprehension, each module generates a detailed textual summary of its operations, including inputs, processes, and outputs. The Command Interpreter compiles these into a complete execution rationale, offering clear insights into the program’s logic and intermediate steps facilitating debugging and refinement. Execution rationales for two representative queries are shown below.

\textit{\textbf{Spatio-temporal Analysis Query:} Perform a trend, seasonality, and neighbourhood analysis of the historical traffic speed data for the past 90 days at location 402117 in the BAY region, focusing on weekdays only. Analyze the Trend Significance, Daily Seasonality Strength, Weekly Seasonality Strength, and Neighbours. The data is recorded at 5-minute intervals and the current time is 2017-03-04 01:40:00.}
\begin{tcolorbox}[
  colback=white,
  colframe=black,
  boxrule=0.5pt,
  arc=0pt,
  left=6pt, right=6pt, top=4pt, bottom=4pt
]
\ttfamily\scriptsize
Loaded data for Location: 402117, Feature: traffic speed, Time Range: From 2016-12-04 01:40:00 to 2017-03-04 01:40:00.\\
Imposed constraints and retrieved data from weekdays only.\\
Trend Analysis Conducted.\\
Trend plot saved at ['./Visualizations/402117\_traffic speed\_trend\_plot.png'].\\
Seasonality Analysis Conducted.\\
Neighbourhood Analysis Conducted.\\
Neighbouring locations detected:[['402056', '402057', '402058', '402118']].\\
Final Explanation Generated.
\end{tcolorbox}
 
\textit{\textbf{Spatio-temporal Prediction Query:} What will be the traffic speed at location 409524 in the BAY region for the next 35 minutes, based on historical data from the past 1 hour recorded at 5-minute intervals? The current time is 2017-02-14 03:00:00. Please ensure that the predicted traffic speed does not exceed 100 km/h. Also, analyze how daily timestamps and neighbouring sensors impact the accuracy of traffic speed predictions.}
\begin{tcolorbox}[
  colback=white,
  colframe=black,
  boxrule=0.5pt,
  arc=0pt,        
  left=6pt, right=6pt, top=4pt, bottom=4pt
]
\ttfamily\scriptsize
Loaded data for Feature: traffic speed, Time Range: From 2017-02-14 02:00:00 to 2017-02-14 03:00:00.\\
Forecasted traffic speed for location 409524 for next 35 minutes using Graph Wavenet Model.\\
Imposed constraints considering traffic speed threshold of 100.\\
Spatial and Temporal Sensitivity analysis conducted.\\
Final Explanation Generated.
\end{tcolorbox}

\section{Experiments} \label{sec:Experiments}
This section presents a comprehensive evaluation of STReason across  key spatio-temporal tasks each requiring varying degrees of data interpretation, inference, and constraint-based reasoning. STReason is benchmarked against several advanced LLM-based baselines to assess its effectiveness in multi-task spatio-temporal inference and execution 
(Section~\ref{sec:MainResults}). We further validate performance through a structured human evaluation that compares the quality and coherence of model-generated responses (Section~\ref{sec:HumanEvaluation}). Additionally, we perform ablation studies to analyze the influence of key 
factors on program generation accuracy, and examine the 
prompt sensitivity and robustness properties of the 
framework (Section~\ref{sec:AblationStudies}).

\subsection{Experimental Setup} \label{sec:Experimental_Setup}
\textbf{Dataset Creation} Due to the lack of standardized datasets for evaluating multi-task spatio-temporal reasoning models, we construct a new dataset tailored to evaluate such capabilities. It encompasses three representative tasks, each 
reflecting distinct spatio-temporal inference challenges. Spatio-temporal Analysis characterizes historical patterns 
through trend, seasonality, and neighbourhood analysis. Spatio-temporal Anomaly Detection identifies statistically unusual observations with temporal and spatial contextualization using auxiliary weather and neighbourhood data. Spatio-temporal Prediction and Reasoning generates forward-looking forecasts with constraint validation and spatio-temporal sensitivity attribution. The dataset comprises 150 structured instances distributed evenly across the three task categories (50 per task), each including a natural language query, a step-by-step ST Program executable by STReason's Command Generator, and a corresponding ground truth answer annotated with key expected components. To ensure broad coverage and generalizability, queries vary across regions, temporal intervals, forecast horizons, and domain-specific constraints. These instances are derived from four diverse real-world datasets including PEMS-BAY and METR-LA \citep{li2017diffusion} for traffic flow, and Beijing\footnote{\url{https://dataverse.harvard.edu/dataverse/whw195009}} and Shenzhen\footnote{\url{https://www.microsoft.com/en-us/research/project/urban-air}} for air quality. This dataset not only enables rigorous evaluation of the STReason framework but also aims to serve as a general-purpose benchmark for the broader research community. 

\textbf{Baselines} We compare STReason framework against six advanced LLM baselines; LLaMA-2-7B\footnote{\url{https://huggingface.co/meta-llama/Llama-2-7b-chat-hf}}, Vicuna-7B v1.5\footnote{\url{https://huggingface.co/lmsys/vicuna-7b-v1.5}}, GPT-3.5 Turbo \citep{ye2023comprehensive}, GPT-4o Mini \footnote{\url{https://platform.openai.com/docs/models/gpt-4o-mini}}, GPT-4 \citep{achiam2023gpt} and DeepSeek-V3 \citep{liu2024deepseek}. These baselines span a range of model sizes and architectural designs, allowing for a comprehensive evaluation across different reasoning capacities and generalization abilities. All models are accessed through public APIs, except for LLaMA-2-7B and Vicuna-7B-v1.5, which are run locally using open-source weights. Table~\ref{tab:baselines} summarizes the key characteristics of all baseline models. 
\begin{table}[h]
\centering
\caption{Summary of Baseline Models}
\label{tab:baselines}
\setlength{\tabcolsep}{3pt}
\small
\begin{tabular}{lccc}
\toprule
\textbf{Model} & \textbf{Parameter Size} & \textbf{Context Length} & 
\textbf{Access Type} \\
\midrule
LLaMA-2-7B      & 7B        & 4k    & Open-source \\
Vicuna-7B-v1.5  & 7B        & 16k   & Open-source \\
GPT-3.5 Turbo   & 175B(est) & 16k   & API \\
GPT-4o Mini     & --        & 128k  & API \\
GPT-4           & 1.7T(est) & 8k    & API \\
DeepSeek-V3     & 671B      & 128k  & API \\
\bottomrule
\end{tabular}
\end{table}

\textbf{Inference Settings}
To ensure a fair comparison across models, we standardized the inference settings for all LLM baselines. For open source models LLaMA-2-7B and Vicuna-7B-v1.5 we used HuggingFace Transformers and answer generation was performed with \texttt{temperature=0.7} and \texttt{max\_new\_tokens=4096}. These two models were run locally using HuggingFace Transformers on a single NVIDIA A40 GPU (46GB memory, CUDA 12.4). The remaining models were accessed via their respective public APIs. For GPT-based models, we used the ChatCompletion endpoint with consistent parameters including \texttt{temperature=0.7}, \texttt{top\_p=1}, and \texttt{max\_tokens=4096}. DeepSeek-V3 was queried using default decoding parameters and \texttt{temperature=1.3} which is used for general conversation. This uniform setup ensured consistent evaluation conditions across all models. 

\textbf{Evaluation Metrics}
To the best of our knowledge, no standardized metrics exist for systematically evaluating long-form question answering and reasoning in spatio-temporal tasks. To address this gap, we propose a novel evaluation framework that jointly measures the correctness, interpretability and reasoning quality of model-generated responses. Informed by the key principles of question answering and reasoning \citep{sun2023survey}, three scores are defined as follows:
\begin{itemize} 
    \item \textbf{Constraint Adherence Score}: Measures whether the generated answer satisfies all query-specified constraints (e.g. temporal granularity, thresholds) using a binary scoring system. An LLM-based verifier receives the query, its associated constraint, and the model response, returning True or False. 
    \item \textbf{Factuality Score}: Assesses the presence and correctness of required analytical components (e.g. detected trends, anomaly timestamps, predicted values) by comparing them against structured ground truth annotations, scored as the ratio of correct components to expected components. 
    \item \textbf{Coherence Score}: Assesses the clarity and logical progression of the explanation on a 3-point ordinal scale: 1 (abrupt or disjointed transitions), 2 (generally clear with minor issues), 3 (seamlessly logical flow). An LLM evaluator assigns scores based on transition quality and reasoning consistency. 
\end{itemize}

In addition to evaluating overall model performance, we also conduct an in-depth assessment of the Program Generator as part of our ablation study (see Section \ref{sec:AblationStudies}). Here, the three metrics Precision, Recall, and F1 Score are computed, considering a program step correct if it matches in module type, input arguments, parameter values, and order. Full details of the evaluation procedure and corresponding prompt formulations are provided in Appendix \ref{app:EvaluationMetrics}.

\subsection{Results} \label{sec:MainResults}
We evaluate STReason across overall performance, qualitative reasoning behavior, forecasting accuracy, cross-domain generalizability and computational efficiency.

\subsubsection{Overall Performance} 
\begin{table*}[htbp]
    \centering
    \caption{Quantitative Comparison of STReason against Baseline Models. Scores are reported as Mean ± Std. The bold and underlined fonts show the best and the second best result respectively.
}
    \setlength{\tabcolsep}{11pt} 

    \begin{subtable}[t]{0.6\textwidth}
        \centering
        \caption{Overall Performance}
        \label{tab:overall_performance}
        \begin{tabular}{|c|c|c|c|}
        \toprule
        \textbf{Model} & \textbf{Constraint} & \textbf{Factuality} & \textbf{Coherence} \\
        & \textbf{Score} & \textbf{Score} & \textbf{Score} \\
        \midrule
        Vicuna-7B-v1.5 & (55.3 \textcolor{black}{± 49.9})\% & (11.6 \textcolor{black}{± 26.4})\% & (71.6 \textcolor{black}{± 35.9})\% \\
        LLaMA-2-7B & (75.3 \textcolor{black}{± 43.3})\% & (12.4 \textcolor{black}{± 23.8})\% & (87.8 \textcolor{black}{± 25.2})\% \\
        GPT-4 & (67.3 \textcolor{black}{± 47.1})\% & (25.1 \textcolor{black}{± 34.9})\% & (99.1 \textcolor{black}{± 7.7})\% \\
        GPT-3.5 Turbo & (88.7 \textcolor{black}{± 31.8})\% & (26.1 \textcolor{black}{± 34.1})\% & (100 \textcolor{black}{± 0.0})\% \\
        GPT-4o Mini & \underline{(98.7 \textcolor{black}{± 11.5})\%} & (29.2 \textcolor{black}{± 38.3})\% & (100 \textcolor{black}{± 0.0})\% \\
        DeepSeek-V3 & (97.3 \textcolor{black}{± 16.2})\% & \underline{(32.8 \textcolor{black}{± 40.4})\%} & (100 \textcolor{black}{± 0.0})\% \\
        STReason & \textbf{(100.0 \textcolor{black}{± 0.0})\%} & \textbf{(84.4 \textcolor{black}{± 26.9})\%} & \textbf{(100 \textcolor{black}{± 0.0})\%} \\
        \bottomrule
        \end{tabular}
    \end{subtable}%
    \hfill
    \begin{subtable}[t]{0.4\textwidth}
        \centering
        \caption{Forecasting Accuracy}
        \label{tab:forecasting_accuracy}
        \begin{tabular}{|c|c|c|}
        \toprule
        \textbf{Model} & \textbf{MAE} & \textbf{RMSE} \\
        \phantom{00.00} & \phantom{00.00} & \phantom{00.00} \\
        \midrule
        Vicuna-7B-v1.5 & 43.1 \textcolor{black}{± 87.6} & 44.1 \textcolor{black}{± 87.5} \\        
        LLaMA-2-7B & 20.9 \textcolor{black}{± 23.2} & 22.4 \textcolor{black}{± 23.7} \\        
        GPT-4 & 52.0 \textcolor{black}{± 19.7} & 52.7 \textcolor{black}{± 19.3} \\        
        GPT-3.5 Turbo & 22.6 \textcolor{black}{± 27.1} & 23.3 \textcolor{black}{± 27.4} \\        
        GPT-4o Mini & \underline{7.9 \textcolor{black}{± 12.3}} & \underline{8.8 \textcolor{black}{± 13.5}} \\ 
        DeepSeek-V3 & 11.4 \textcolor{black}{± 18.0} & 12.4 \textcolor{black}{± 18.7} \\        
        STReason & \textbf{7.6 \textcolor{black}{± 11.9}} & \textbf{8.4 \textcolor{black}{± 12.8}} \\

        \bottomrule
        \end{tabular}
    \end{subtable}
\end{table*}
Table \ref{tab:overall_performance} presents a comparative evaluation of STReason against six advanced LLM baselines across three evaluation metrics constraint adherence, factuality, and coherence. The results clearly demonstrate the superiority of STReason in effectively handling complex spatio-temporal tasks requiring both computational precision and interpretability.  Specifically, STReason achieves a perfect constraint adherence score, satisfying all task-specific requirements across queries. While strong baselines such as GPT-4o Mini and DeepSeek-V3 perform competitively, STReason is the only framework that consistently meets all constraints without exception. The most pronounced improvement is observed in the factuality score, where STReason achieves 84.44\%, demonstrating its ability to extract relevant analytical components and produce factually correct responses. With regard to the coherence score, along with models like GPT-3.5, GPT-4o Mini, and DeepSeek, STReason delivers a top performance of 100\%, maintaining complete logical consistency in its explanations. This reflects the inherent strengths of advanced LLMs in generating fluent and well-structured language. Statistical significance tests further confirm that STReason’s improvements in constraint adherence and factuality are significant compared to the majority of baselines (Appendix~\ref{app:StatisticalSignificance}). Overall, STReason’s strong performance across all dimensions highlights its reliable internal mechanisms and task relevant knowledge to generate coherent and factually accurate outputs. These results further establish STReason as a powerful framework for long-form, multi-task spatio-temporal reasoning tasks. 

\subsubsection{Qualitative Reasoning Behavior}
Beyond quantitative metrics, we observed systematic qualitative differences in model outputs across five distinct failure categories, contrasted with STReason's detailed, specific and statistically grounded responses. Table~\ref{tab:qualitative_comparison} summarizes 
these patterns with representative examples.

\begin{table}[h]
\centering
\caption{Qualitative Output Comparison across Models}
\label{tab:qualitative_comparison}
\setlength{\tabcolsep}{3pt}
\small
\begin{tabular}{p{1.4cm}|p{1.8cm}|p{4.3cm}}
\toprule
\textbf{Model} & \textbf{Failure Type} & 
\textbf{Representative Behavior} \\
\midrule
Vicuna-7B & Hallucination & Fabricates seasonal trends 
across summer/winter months despite receiving only 30 days 
of data; repeats query as a list of instructions with no 
computation. \\
\midrule
LLaMA-2-7B & Irrelevant steps & Outputs generic procedural 
steps (e.g., ``apply ARIMA", ``perform clustering") 
unrelated to the query; produces no computed results. \\
\midrule
GPT-3.5 / GPT-4 & High-level explanation without execution & 
Provides methodological descriptions without executing 
analyses. GPT-4 explicitly states ``As an AI, I can't 
dynamically predict data", then offers a rough static 
estimate. No constraint validation performed. \\
\midrule
GPT-4o Mini / DeepSeek-V3 & Trivial forecasts & Produces 
structured steps but defaults to persistence or 
average-based forecasting; applies no advanced model, 
lacks constraint enforcement, and provides no sensitivity attribution. \\
\bottomrule
\end{tabular}
\end{table}

These observations highlight a clear capability gap where baseline models either fabricate patterns from insufficient data, describe methods without executing them, or rely on trivial forecasting heuristics. STReason's execution-grounded pipeline systematically avoids these failure modes by constraining generation to verified computational outputs, delivering comprehensive, precise, and actionable responses for complex spatio-temporal tasks. Full output examples for each model are provided in Appendix~\ref{app:extended_appendix}.

\subsubsection{Forecasting Accuracy}
\sloppy
We further conducted a forecasting accuracy comparison using Mean Absolute Error (MAE) and Root Mean Squared Error (RMSE), the standard metrics in spatio-temporal forecasting (Table \ref{tab:forecasting_accuracy}). Unlike the baselines, which often produced incomplete or missing prediction sequences, STReason consistently generated complete, correctly formatted outputs. For fair comparison, baseline predictions were post processed by zero padding if no predictions were generated and repeating the last available value to complete partial prediction sequences. As shown in Table \ref{tab:forecasting_accuracy}, STReason achieves the lowest MAE and RMSE, demonstrating superior forecasting accuracy. Moreover, these forecasting improvements are statistically significant against most baseline models, further confirming its superior quantitative reliability (see Appendix \ref{app:StatisticalSignificance}). Although models like GPT-4o Mini and DeepSeek showed competitive results, the post-processing adjustments could have unfairly favored these models by smoothing missing values, potentially inflating their apparent  performance. Despite this, STReason still achieves the best accuracy, confirming its strength in delivering both accurate quantitative predictions and logically sound reasoning outputs.

\subsubsection{Cross-domain Generalizability}
We also assessed the generalizability of STReason by applying it to two unseen domains: epidemiology and finance without modifying the core architecture or prompting strategy. The only adjustments involved were updating the data retrieval function to point to the relevant sources, and replacing the forecasting module with TimesNet \citep{wu2022timesnet} for pure time-series inputs. The following epidemiology and finance examples illustrate STReason’s ability to perform meaningful spatio-temporal analysis and prediction in domains beyond traffic and air quality. These results highlight the flexibility of STReason's modular design. Across both domains, STReason generated contextually grounded explanations that correctly interpreted domain-specific terminology (e.g., ILI case counts, financial stability thresholds) without any domain-specific prompt engineering, demonstrating the framework's adaptability to unseen domains.
\vspace{6pt}

\mdfdefinestyle{examplebox}{
  linecolor=black!40,
  linewidth=0.3pt,
  backgroundcolor=gray!5,
  innerleftmargin=4pt,
  innerrightmargin=4pt,
  innertopmargin=4pt,
  innerbottommargin=4pt,
  frametitlebackgroundcolor=gray!20,
  frametitlefont=\bfseries\small,
  frametitlerule=true,
  frametitlerulecolor=black!40,
  frametitlerulewidth=0.3pt,
  frametitleaboveskip=3pt,
  frametitlebelowskip=3pt
}
\begin{mdframed}[style=examplebox, frametitle={Epidemiology Example}]

\textbf{Query:} Perform a seasonality analysis of historical national ILI case counts for the past 52 weeks. The current time is 2019-01-22 00:00:00. The data is recorded at every week. Analyze the Monthly Seasonality Strength.

\smallskip
\textbf{Generated Answer:} The data represents the number of ILI (Influenza-Like Illness) case counts recorded every week. The data starts with 23169 cases in the first week and peaks at 70056 cases in the 15th week before gradually decreasing to 27419 cases in the last week. The monthly seasonality strength is calculated to be 0.00036, indicating a very weak monthly seasonality pattern in the data. This suggests that the fluctuations in the number of ILI case counts do not follow a consistent monthly pattern and are more likely influenced by other factors. The low monthly seasonality strength suggests that the variations in the number of ILI case counts are not significantly affected by the month in which the data was collected. This could be due to factors such as changes in weather, vaccination campaigns, or other external factors that impact the spread of influenza-like illnesses.
\end{mdframed}
\vspace{3pt}

\begin{mdframed}[style=examplebox, frametitle={Finance Example}]
\textbf{Query:} The current time is 2010-07-06 12:00:00. Predict the exchange rate of GBP to USD for the next 5 days using the historical data of the past 96 days with data points recorded at daily intervals. Ensure predictions stay below the 1.55 financial stability threshold. Analyze how temporal patterns affect the forecast.

\smallskip
\textbf{Generated Answer:} The predicted exchange rate values for the next 5 days are as follows:
\[
[1.3212, \; 1.3242, \; 1.3186, \; 1.3189, \; 1.3200].
\]
These predictions were generated using the deep learning-based TimesNet Model. All predictions fall within the exchange rate threshold, indicating that the model's forecasts are within acceptable limits and align with historical data. The timestamp sensitivity scores provide insights into the temporal factors influencing the exchange rate predictions. Among the timestamps analyzed, the strongest influences were observed on July 4, 2010, July 5, 2010, and July 6, 2010, with impacts of 0.19, 0.17, and 0.17 respectively. These dates likely correspond to significant events or announcements that had a notable impact on the exchange rate forecast. By analyzing these timestamps in conjunction with macroeconomic trends, we can gain a deeper understanding of the drivers behind the predicted exchange rate values.
\end{mdframed}

\subsubsection{Latency and Cost Analysis}
To evaluate the practical feasibility of deploying STReason in real world decision-making settings, we finally analyzed its latency, token usage and API cost across representative query types. All evaluations were conducted using GPT-3.5 Turbo on standard hardware without GPU acceleration.
\begin{table}[htbp]
    \centering
    \caption{Average Latency by Task Type (in seconds)}
    \label{tab:latency_by_task}
    \setlength{\tabcolsep}{4pt}
    \small
    \begin{tabular}{lccc}
    \toprule
    \textbf{Task Type} & \textbf{Program Gen} & \textbf{Answer Gen} & \textbf{Total Time} \\
    \midrule
    Analysis                      & 4.27 & 5.04  & 9.31  \\
    Anomaly Detection             & 4.20 & 6.92  & 11.12 \\
    Prediction and Reasoning      & 4.59 & 10.91 & 15.50 \\
    \bottomrule
    \end{tabular}
\end{table}

Table \ref{tab:latency_by_task} reports the average latency results for three task categories. Each query goes through two LLM calls, one for Program Generation and one for Answer Generation, including function execution and explanation synthesis. Analysis tasks are completed in approximately 9 seconds, demonstrating rapid end-to-end execution. Anomaly detection, which requires deeper contextual reasoning, takes slightly longer. Prediction tasks involve generating multi-step predictions and explanatory narratives, resulting in the highest average latency of approximately 16 seconds. Overall, all task types remain within an acceptable response window for interactive applications, confirming the practical feasibility of deploying STReason in urban monitoring and planning systems.

Token usage in STReason varies with the complexity and task type of the query. On average, queries are $\sim80$ tokens, and responses are $\sim364$ tokens. In addition, the input includes a structured Function Pool ($\sim2211$ tokens) and in-context examples ($\sim700$ tokens). This results in a total input size of  $\sim(2800-3000)$ tokens per query, which remains well within the 16k context limit of GPT-3.5 Turbo. At OpenAI's pricing for GPT-3.5 Turbo at time of evaluation (\$0.0015 per 1K input tokens and \$0.002 per 1K output tokens), the estimated cost per query is approximately \$0.0051, making STReason highly cost-efficient for scalable deployment.



\subsection{Human Evaluation Experiment} \label{sec:HumanEvaluation}
To further validate the performance of STReason and assess the credibility, clarity, and reasoning quality of model generated answers we conducted a rigorous human evaluation. This complements our quantitative results with human judgment, thereby offering deeper insights into the effectiveness of STReason in real world contexts. We recruited 27 evaluators with domain relevant expertise, nearly half holding Ph.D.s and the majority specializing in Computing, Statistics, or Data Science, ensuring a technically competent cohort (see Appendix \ref{app:extended_appendix}). For this study, 18 spatio-temporal queries were curated spanning three core task types targeted by STReason. Each query was paired with two answers, one from STReason and one from a randomly selected baseline, with each baseline compared exactly three times. The order of answers was randomized to avoid positional bias. Evaluators were asked to choose the better response and provide open-ended feedback to explain their preferences. Full evaluation details are provided in Appendix \ref{app:extended_appendix}.
\begin{figure}[htbp]
    \centering
    \begin{minipage}{0.45\textwidth}
        \includegraphics[width=\linewidth]{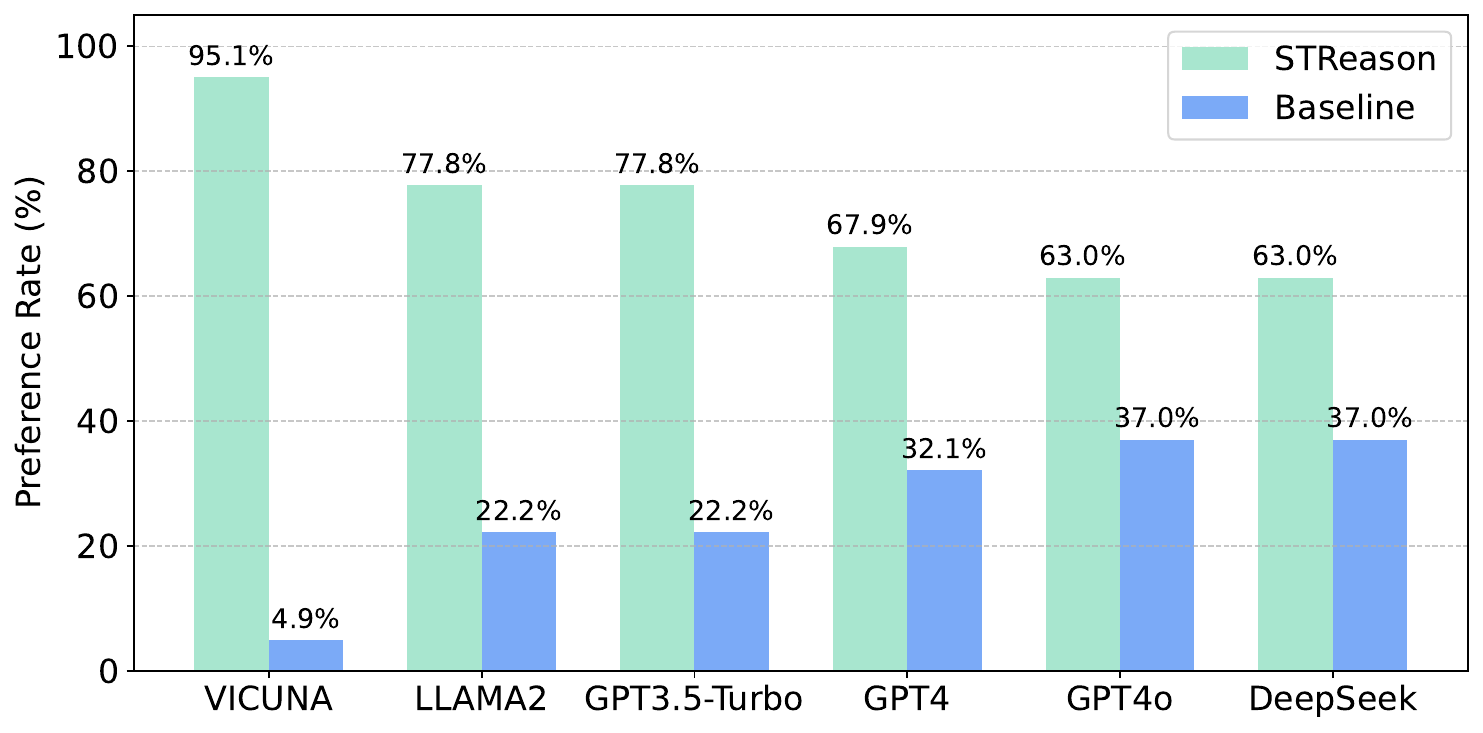}
        \caption{STReason vs. Baseline Preference Rate}
        \label{fig:HumanEvalBaseline}
    \end{minipage}
    \hfill
    \begin{minipage}{0.45\textwidth}
        \centering   
        \includegraphics[width=\linewidth]{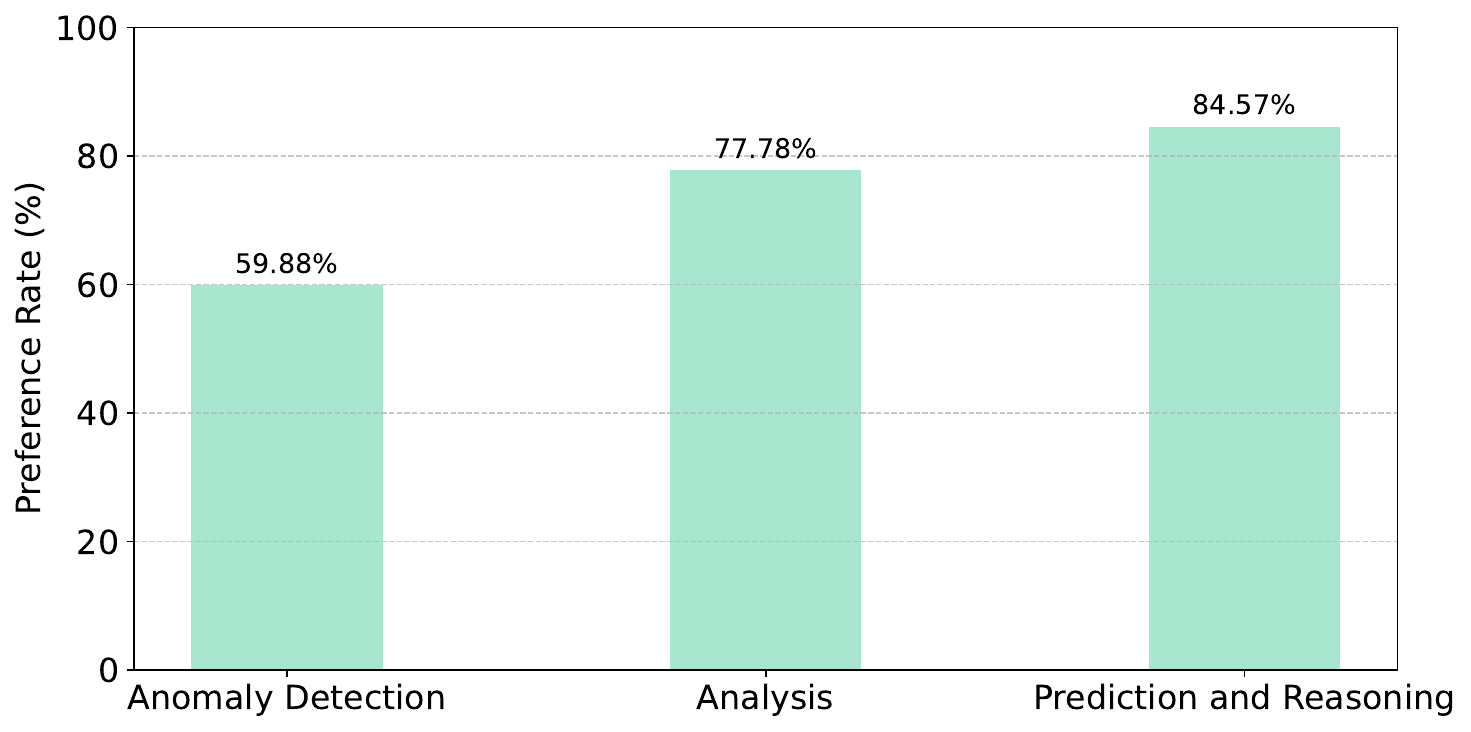}
        \caption{STReason Task-wise Preference Rate}
        \label{fig:HumanEvalTask}
    \end{minipage}
\end{figure}

Across all 486 comparisons (27 evaluators × 18 questions), STReason was preferred in 74.1\% of responses on average demonstrating strong user preference and credibility of generated outputs. As shown in Figure \ref{fig:HumanEvalBaseline}, STReason consistently outperformed all baselines. Vicuna-7B exhibited the lowest preference rate, followed by LLaMA-2-7B, GPT-3.5 Turbo and GPT-4 respectively. Even against top competitors like GPT-4o Mini and DeepSeek, STReason maintained a clear margin of superiority. Notably, there is a  consistent alignment between human evaluator preferences (Figure \ref{fig:HumanEvalBaseline}) and the automated factuality scores (Table \ref{tab:overall_performance}). The order of preference remains similar in both evaluations, which highlights the robustness and reliability of the proposed evaluation framework. Furthermore, as illustrated in Figure \ref{fig:HumanEvalTask}, the task wise preference rate further reveals STReason's robust performance across spatio-temporal task categories. Specifically it achieves 84.57\% preference rate in the prediction and reasoning task, 77.78\% in analysis, and 59.88\% in anomaly detection. 

These results suggest that while the model excels generally across all tasks, there is more room for improvement in tasks involving rare event detection. Three factors 
contribute to its relatively lower performance. First, 
the \texttt{DETECT\_ANOMALY\_ST\_DATA} module requires joint 
reasoning over three data sources simultaneously, including primary data, temporal auxiliary weather data, and spatial 
auxiliary neighbourhood data making it the most 
compositionally complex ST Program in the framework. This increases the surface area for program 
generation errors, consistent with the precision dip 
observed in Figure~\ref{fig:Ablation_task}. Second, 
genuine anomalies are rare within typical 7-30 day 
query windows, creating class imbalance that inflates 
false positive rates in the detection module. Third, 
anomaly detection task is inherently more subjective 
than trend or prediction tasks, introducing greater 
variance in ground truth annotation and human evaluation 
judgments. Addressing these limitations through 
uncertainty-aware anomaly scoring and adaptive threshold 
calibration remains an avenue for future work.

\begin{figure}[htbp]
    \centering
    \includegraphics[width=\linewidth]{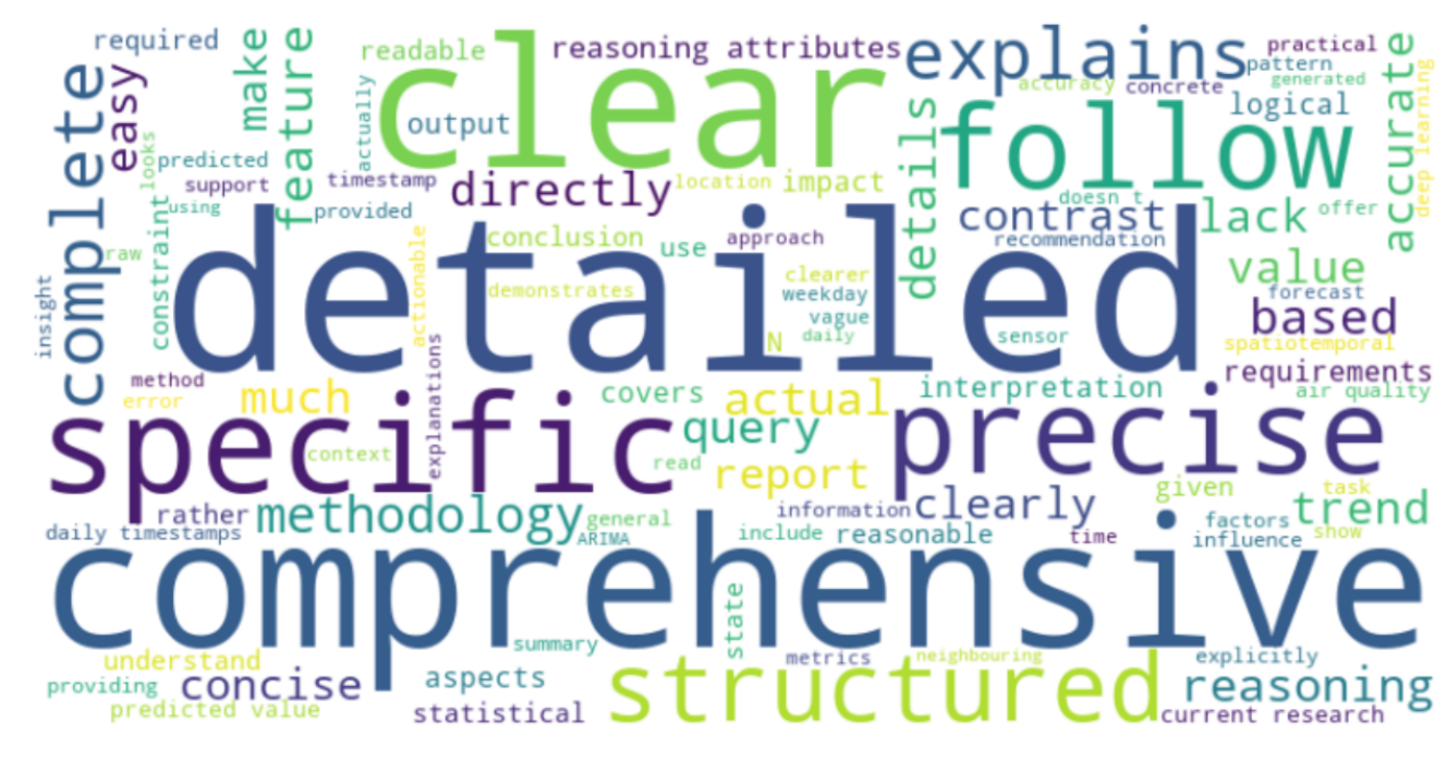}
    \caption{Human Evaluation: Qualitative Feedback}
    \label{fig:HumanEval_WordCloud}
\end{figure}

Moreover, qualitative feedback from evaluators further supported these findings. As shown in Figure \ref{fig:HumanEval_WordCloud}, the word cloud generated from open-ended responses where STReason was preferred, highlights attributes such as “detailed”, “comprehensive”, “clear”, “structured”, and “precise” reflecting STReason’s strength in delivering well-organized, transparent, and informative answers. Overall, these outcomes affirm STReason’s capacity to not only outperform baselines in constraint adherence and factual accuracy, but also to produce human preferred outputs that are coherent, more interpretable and contextually grounded.

\subsection{Ablation Studies} \label{sec:AblationStudies}
To better understand the contribution of core components within the STReason framework, we conduct a series of ablation studies centered on the Command Generator. Each experiment is evaluated against ground-truth programs using Precision, Recall, and F1 Score as defined in Section \ref{sec:Experimental_Setup}.

\textbf{Effect of In-context Example Variant:} We first investigate how different in-context example variants influence program correctness. Four configurations were tested; a)\,\textit{Equal Construction:} A fixed and balanced set of task-specific examples, b)\,\textit{Random Construction:} A set of randomly selected examples from the example pool, c)\,\textit{Test-Query Include:} A set of randomly selected examples including a certain percentage (20\% in our case) of examples similar to the test query, c)\,\textit{Test-Query Exclude:} A set of randomly selected examples excluding examples similar to the test query. As shown in Table \ref{tab:incontext_variants}, the Equal Construction yields the highest performance, affirming the importance of balanced, task-aligned examples. While the Test-Query Include setup also performs strongly, the Random setup demonstrates a noticeable drop in precision due to decreased example task alignment. The Test-Query Exclude configuration exhibits the lowest performance across metrics, highlighting the challenge of generating correct programs when the example pool lacks task relevance to the query. 
\begin{table}[h]
\centering
\caption{Effect of In-context Example Variant}
\label{tab:incontext_variants}
\begin{tabular}{l|c|c|c|c}
\toprule
\textbf{} & \textbf{Equal} & \textbf{Random} & \textbf{Test-query} & \textbf{Test-query} \\ 
\textbf{} & \textbf{} & \textbf{} & \textbf{Include} & \textbf{Exclude} \\ 
\midrule
\textbf{Precision} & \textbf{0.9816} & 0.8327 & 0.9642 & 0.6091 \\
\textbf{Recall} & \textbf{1.0000} & 0.9973 & 1.0000 & 1.0000 \\
\textbf{F1 Score} & \textbf{0.9907} & 0.9026 & 0.9818 & 0.7571 \\
\bottomrule
\end{tabular}
\end{table}

\begin{table*}[h!]
\centering
\caption{Effect of Function Pool}
\label{tab:Effect_FuncPool}
\resizebox{\textwidth}{!}{
\begin{tabular}{l|c|c|c|c|c|c|c|c}
\toprule
\multirow{2}{*}{} & \multicolumn{4}{c|}{\textbf{Precision}} & \multicolumn{4}{c}{\textbf{F1 Score}} \\
\cmidrule{2-9}
 & \textbf{Equal} & \textbf{Random} & \textbf{Test-Query} & \textbf{Test-query} & \textbf{Equal} & \textbf{Random} & \textbf{Test-Query} & \textbf{Test-Query} \\
 & \textbf{} & \textbf{} & \textbf{Include} & \textbf{Exclude} & \textbf{} & \textbf{} & \textbf{Include} & \textbf{Exclude} \\
\midrule
\textbf{W/O Function Pool} & 0.9816 & 0.8327 & 0.9642 & 0.6091 &  0.9907 & 0.9026 & 0.9818 & 0.7571 \\
\textbf{With Function Pool} & \textbf{0.9874} & \textbf{0.9678} & \textbf{0.9687} & \textbf{0.8440} & \textbf{0.9936} & \textbf{0.9836} & \textbf{0.9841} & \textbf{0.9132} \\
\bottomrule
\end{tabular}
}
\end{table*}

\textbf{Effect of Function Pool:} We further evaluate the benefit of augmenting examples with a curated function pool. As shown in Table \ref{tab:Effect_FuncPool}, the presence of the function pool significantly boosts precision and F1 scores across all example construction settings. It is particularly beneficial in the Test-Query Exclude setting, where query relevant examples are absent. These findings highlight the value of providing structured functional knowledge alongside in-context examples in improving program generation accuracy, especially when task-specific cues in the examples are limited. In summary, Tables~\ref{tab:incontext_variants} 
and~\ref{tab:Effect_FuncPool} collectively serve as a 
prompt sensitivity analysis for STReason. The four 
in-context configurations represent systematically 
varying levels of prompt distribution shift, from 
optimal (Equal Construction) to adversarial (Test-Query 
Exclude). Even under the most challenging configuration, 
STReason retains an F1 Score of 0.9132 with the Function 
Pool, representing a drop of only 8.0\% from the optimal 
setting. This demonstrates that STReason maintains 
meaningful program generation accuracy under significant 
prompt variation, with the Function Pool acting as a 
structural robustness mechanism that compensates for 
reduced example relevance.

\begin{figure}[htbp]
    \centering
    \begin{minipage}{0.45\textwidth}
        \includegraphics[width=\linewidth]{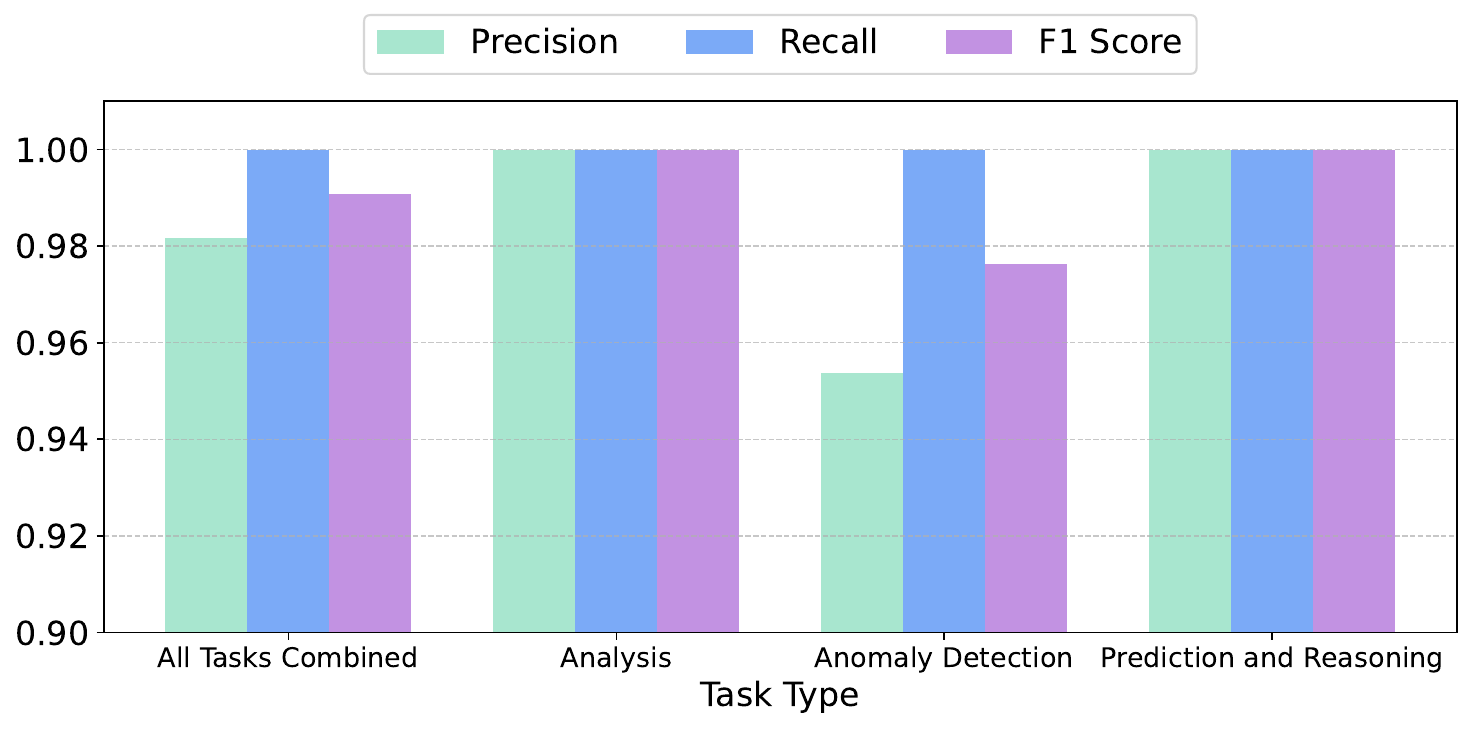}
        \caption{Effect of Task}
        \label{fig:Ablation_task}
    \end{minipage}
    \hfill
    \begin{minipage}{0.45\textwidth}
        \centering   
        \includegraphics[width=\linewidth]{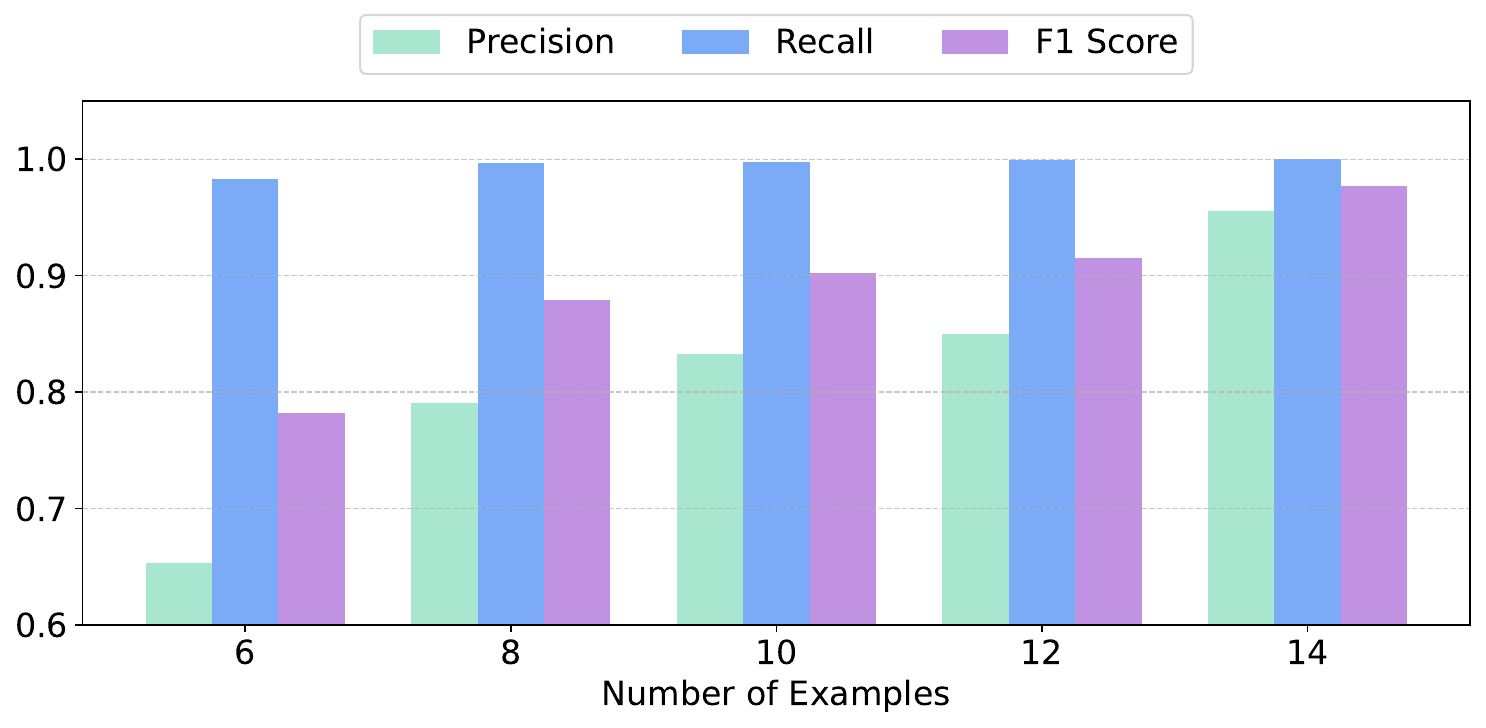}
        \caption{Effect of No: of Incontext Examples}
        \label{fig:Ablation_examples}
    \end{minipage}
\end{figure}
\textbf{Effect of Task Type:} We then assess performance across different task categories. As shown in Figure \ref{fig:Ablation_task}, the generator achieves perfect scores in both Analysis and Prediction tasks, while Anomaly Detection proves more challenging, showing a dip in precision, potentially due to the variability in task-specific steps and parameter configurations. However, when considering all tasks combined, the command generator maintains high overall accuracy, demonstrating strong generalizability.

\textbf{Effect of Number of In-Context Examples:} Next, we also evaluate the impact of varying the number of in-context examples on program generation accuracy (see Figure \ref{fig:Ablation_examples}). Precision and F1 Score steadily improve as the number of examples increases from 6 to 14, while recall remains consistently high across all configurations. This trend highlights the importance of sufficient context in helping the model generate more accurate programs. Furthermore, it is observed that, as the number of in-context examples increases, the proportion of task specific examples within the example pool rises from 14\% with 6 examples to 21\% with 14 examples contributing to the performance gains.


\section{Conclusion, Limitations and Future Work} \label{sec:Conclusion, Limitations and Future Work}
In this work, we introduced STReason, a novel training-free 
framework for spatio-temporal multi-task inference and reasoning 
that integrates the reasoning strengths of large language models 
with the analytical capabilities of state-of-the-art 
spatio-temporal models. Through extensive benchmarking we show that STReason outperforms advanced LLM baselines in spatio-temporal inference and execution tasks, achieving superior constraint adherence, factual accuracy, and reasoning coherence. Notably, our findings are further validated through structured human evaluations, with human evaluator preferences aligning consistently with automated factuality scores, confirming the reliability of the proposed evaluation framework. Beyond the framework itself, we introduce a new benchmark dataset for long-form multi-task spatio-temporal reasoning, alongside a unified evaluation framework designed to serve the broader research community. 

While effective, STReason currently relies on manually curated in-context examples which may limit scalability to unseen task types without further adaptation. Additionally, performance on tasks such as anomaly detection remains an area for improvement. Future work will focus on automating example retrieval and function pool maintenance to further enhance generalization, and expanding task coverage to include more complex query types across diverse domains. Additionally, validating STReason across multiple LLM backbones, including newer frontier models remains an important direction for establishing the generalizability of the framework beyond its current backbone. We also plan to extend STReason into a unified reasoning agent that can operate interactively across multimodal spatio-temporal environments.

\clearpage
\bibliographystyle{ACM-Reference-Format}
\bibliography{References}

\clearpage
\appendix
\section{Appendix}
\subsection{Extended Comparison: Program-based Reasoning 
Frameworks} \label{app:related_work_ext}
Table~\ref{tab:rel_work_program} contrasts STReason with 
representative program-based reasoning frameworks across 
domain, input modality, output format, and key innovation. While these frameworks share the principle of decomposing natural language queries into structured programs, direct empirical comparisons with STReason are not feasible due to fundamental differences in problem formulation, data representation, and evaluation objectives. For example, VISPROG processes visual inputs and handles questions such as \textit{``How many muffins can each kid have in the picture?"}, whereas STReason is designed for analytical spatio-temporal queries such as \textit{``Perform a trend, seasonality, and neighborhood analysis of the historical traffic speed data for the past 90 days at location 402117 in the BAY region."}.

\begin{table*}[htbp]
\centering
\caption{Comparison of STReason with representative program-based reasoning frameworks.} 
\label{tab:rel_work_program}
\scriptsize
\renewcommand{\arraystretch}{1.4}
\setlength{\tabcolsep}{4pt}
\begin{tabular}{|>{\raggedright\arraybackslash}m{2.2cm}|
                >{\raggedright\arraybackslash}m{2.2cm}|
                >{\raggedright\arraybackslash}m{2.2cm}|
                >{\raggedright\arraybackslash}m{4.0cm}|
                >{\raggedright\arraybackslash}m{4.0cm}|}
\hline
\textbf{Method} & \textbf{Domain} & \textbf{Input Modality} & \textbf{Output Format} & \textbf{Key Innovation} \\
\hline
VISPROG & Vision QA & Images & Short categorical text/ Edited Images & Compositional visual reasoning programs \\
\hline
ViperGPT & Vision QA & Images & Short categorical text/ Edited Images & Integrating code-generation models into vision \\
\hline
Chain-of-Table & Tabular reasoning & Structured tables & Short categorical text/ Result Table & Reasoning chains over evolving tables \\
\hline
HuggingGPT & General AI services & Multimodal (text, image, audio) & Task-specific responses & LLM routes tasks to expert models \\
\hline
Program-of-Thoughts & Math/text QA & Text & Numeric or symbolic answers & Program synthesis for mathematical reasoning \\
\hline
UrbanLLM & Spatio-temporal QA (Urban Planning) & Location and Time indexed urban data & Numeric Forecasts/Short answers & Domain-tuned LLMs for traffic/urban tasks \\
\hline
\textbf{STReason (ours)} & \textbf{Spatio-temporal QA (Multi-Domain)} & \textbf{Location and Time indexed data} & \textbf{Numeric analysis + Long-form Reasoning + Execution Rationale}  & \textbf{Analytical, constraint-aware execution for interpretable, multi-domain ST reasoning }\\
\hline
\end{tabular}
\end{table*}

To further highlight why comparisons with existing program-based methods are infeasible, we tested representative STReason queries on several open-source program-based frameworks. Chain-of-Table returned \textit{``Final answer: N/A"}, Program-of-Thoughts failed with \textit{``Empty solver() function and Prediction: None"}, and PAL (Program-aided Language Models) returned \textit{``No results was produced. A common reason is that the generated code snippet is not valid or did not return any results."}. These outcomes show that, although effective in their respective domains, existing models are unable to process the structured spatio-temporal queries that STReason is designed to address. STReason therefore advances beyond both general program-based frameworks and domain-specific approaches such as UrbanLLM, introducing mechanisms for structured, actionable reasoning in spatio-temporal decision-making. In summary, STReason is a deliberate extension of program-guided reasoning to real-world problems requiring structured inputs, statistical analysis, and interpretable outputs, thereby filling a critical gap between LLMs and decision-support tools for spatio-temporal domains.

\subsection{Task-wise Program Generation }\label{app:ProgramGen_Tasks}
\begin{figure}[htbp]
    \centering
    \includegraphics[width=\linewidth]{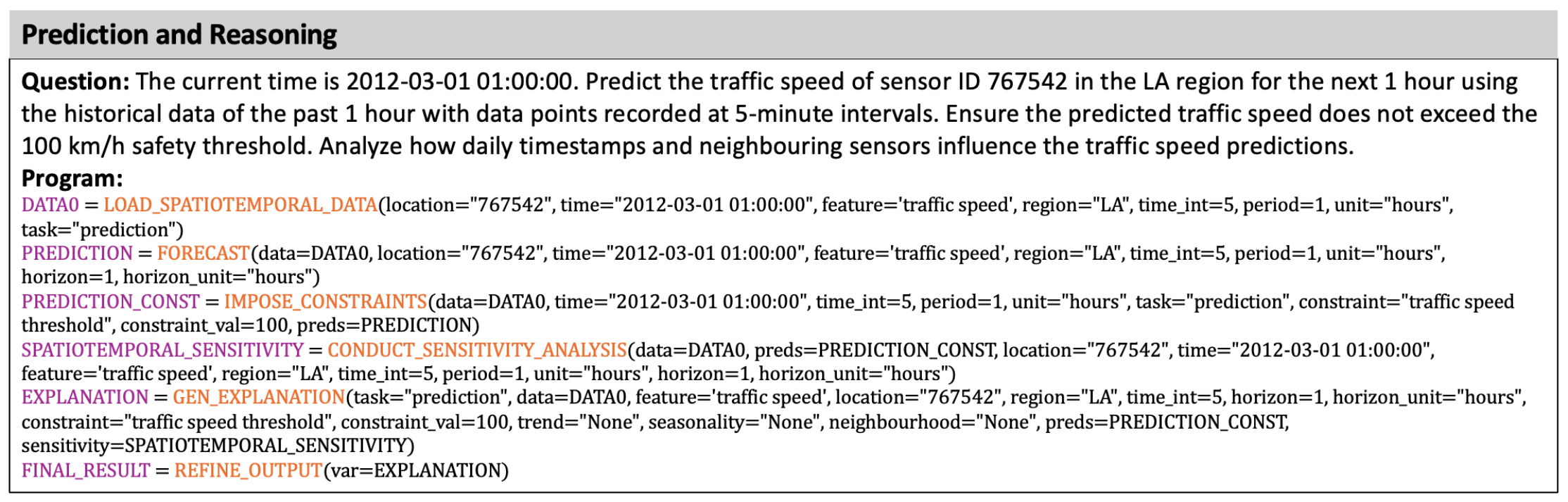}
    \caption{ST-program for Prediction and Reasoning Task}
    \label{fig:ProgramGen_Prediction}
\end{figure}

\begin{figure}[htbp]
    \centering
    \includegraphics[width=\linewidth]{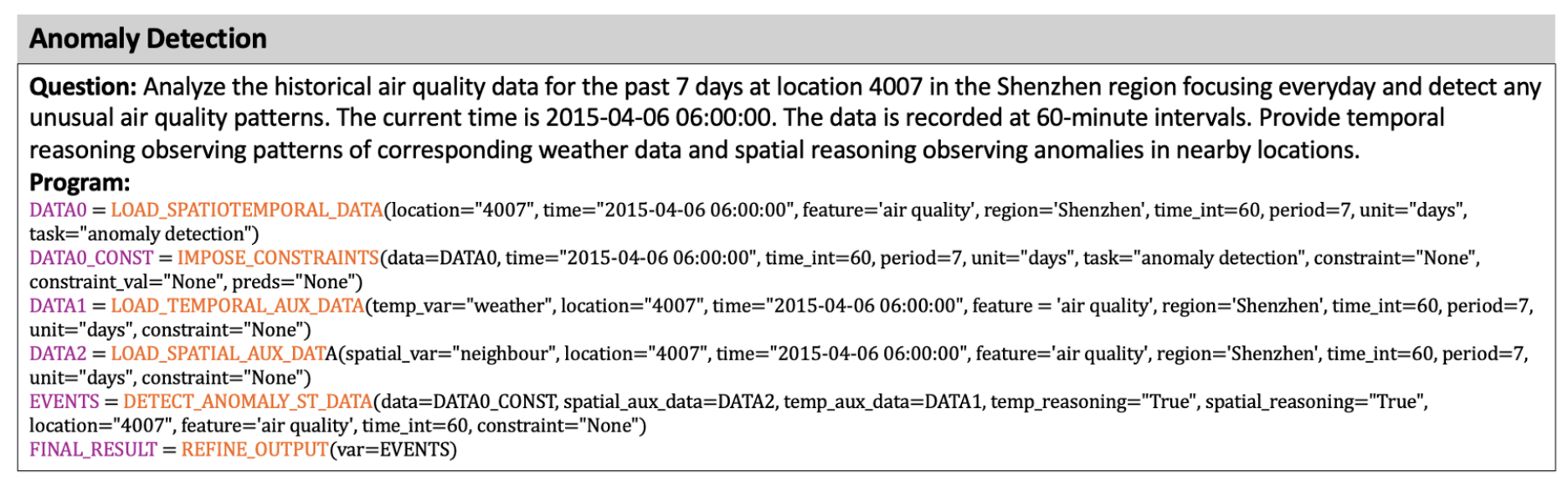}
    \caption{ST-program for Anomaly Detection Task}
    \label{fig:ProgramGen_Anomaly}
\end{figure}

We showcase sample ST-program structures generated for the spatio-temporal Prediction and Reasoning task in Figure~\ref{fig:ProgramGen_Prediction} and the spatio-temporal Anomaly Detection task in 
Figure~\ref{fig:ProgramGen_Anomaly}.

\subsection{Evaluation Metrics}\label{app:EvaluationMetrics}
Prompt formulations for the proposed evaluation metrics are shown in Figures \ref{fig:constraint_prompt},\ref{fig:factuality_prompt} and \ref{fig:coherence_prompt}. We also assess the accuracy of the ST-Program generated by the Command Generator, which form the backbone of STReason’s execution process. A program step is considered correct if it matches the reference in module type, argument names, parameter values, and order. To measure program generation performance, we adopt the following metrics:

\begin{figure*}[htbp]
    \centering
    \begin{minipage}{0.32\textwidth}
        \includegraphics[width=\linewidth]
        {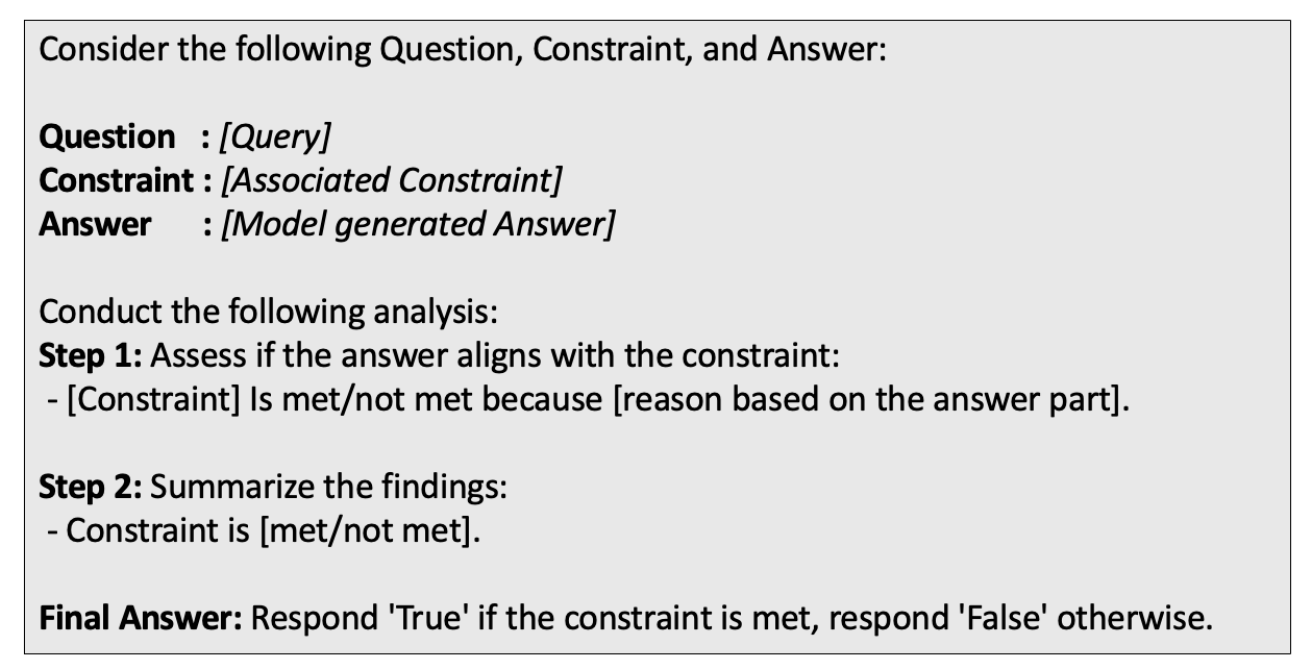}
        \caption{Constraint Adherence Prompt}
        \label{fig:constraint_prompt}
    \end{minipage}
    \hfill
    \begin{minipage}{0.32\textwidth}
        \includegraphics[width=\linewidth]
        {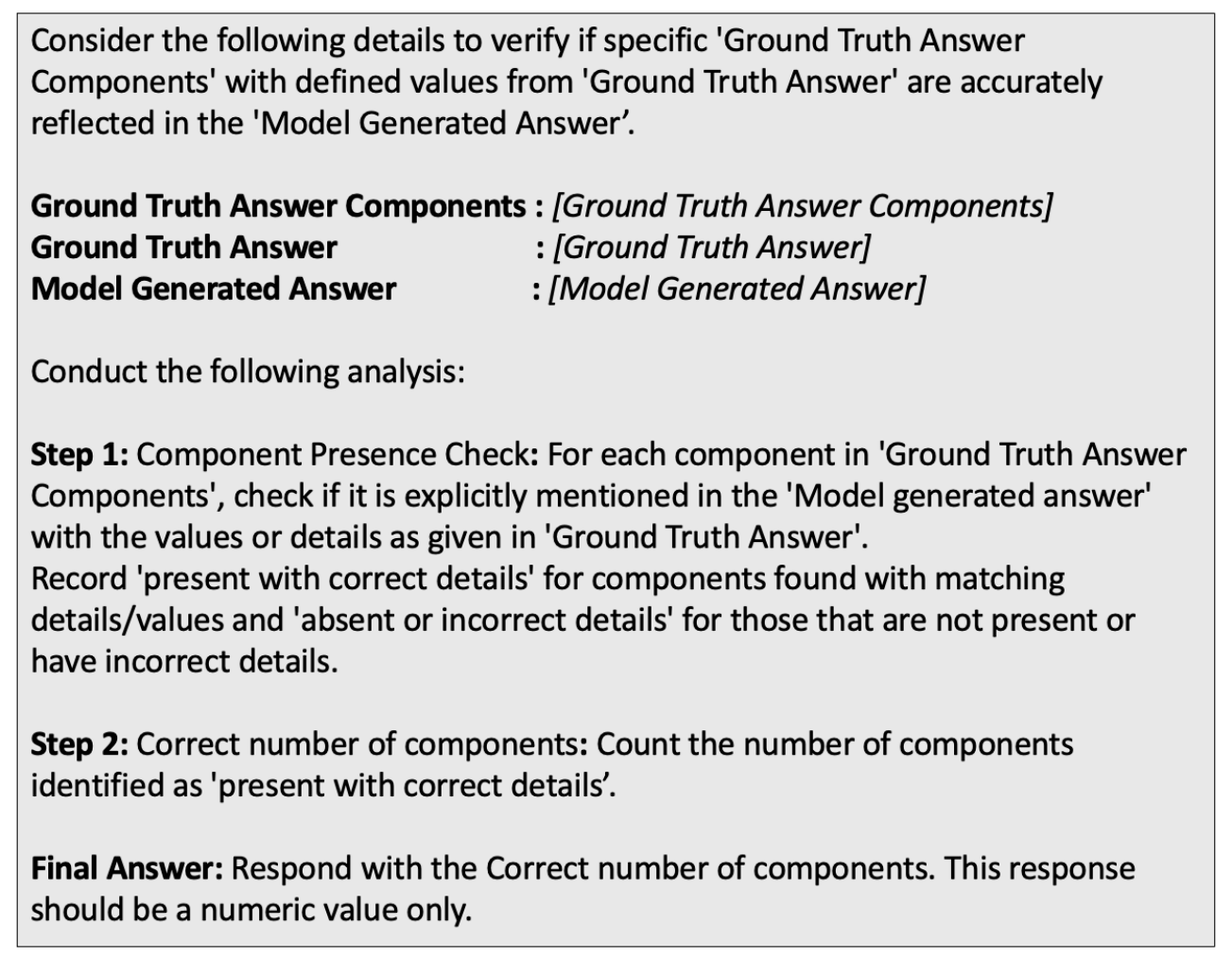}
        \caption{Factuality Score Prompt}
        \label{fig:factuality_prompt}
    \end{minipage}
    \hfill
    \begin{minipage}{0.32\textwidth}
        \includegraphics[width=\linewidth]
        {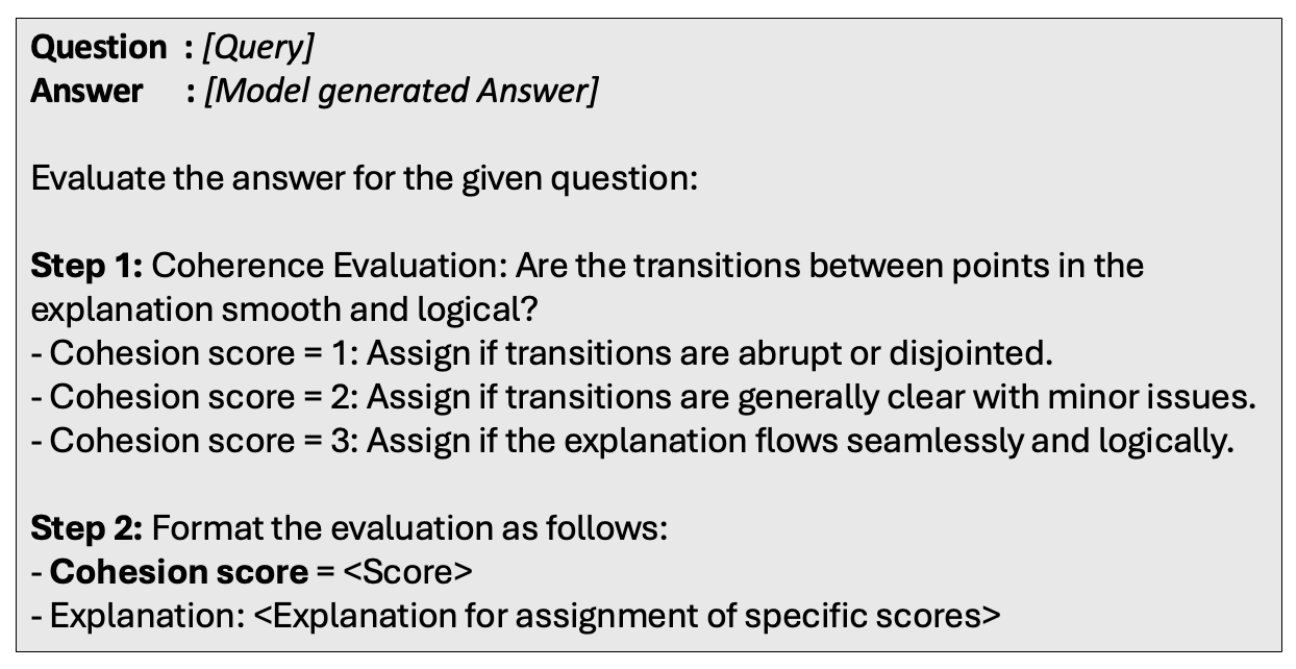}
        \caption{Coherence Score Prompt}
        \label{fig:coherence_prompt}
    \end{minipage}
\end{figure*}

\textbf{Precision} evaluates the accuracy of predicted steps by measuring the proportion of generated commands that are both syntactically and semantically correct. 
\begin{equation}
\text{Precision} = \frac{\text{True Positives (TP)}}{\text{True Positives (TP)} + \text{False Positives (FP)}}
\end{equation}

\textbf{Recall} measures the completeness of the generated program by calculating the proportion of required steps that the model successfully included. 
\begin{equation}
\text{Recall} = \frac{\text{True Positives (TP)}}{\text{True Positives (TP)} + \text{False Negatives (FN)}}
\end{equation}

\textbf{F1-score} provides a balanced measure by computing the harmonic mean of precision and recall, offering a single metric to capture the trade-off between completeness and correctness:
\begin{equation}
\text{F1 Score} = 2 \times \left( \frac{\text{Precision} \times \text{Recall}}{\text{Precision} + \text{Recall}} \right)
\end{equation}

\begin{itemize}
    \item True Positives (TP): Program steps that exactly match the ground truth in module type, argument names, and values.
    \item False Positives (FP): Steps generated by the model that either do not appear in the ground truth, or are incorrect in terms of function, parameters, or sequence.
    \item False Negatives (FN): Steps present in the ground truth but missing from the generated program, such as omitted data loading or post-processing functions.
\end{itemize}


\subsection{Statistical Significance Testing} \label{app:StatisticalSignificance}
\textcolor{black}{To assess the robustness of STReason’s performance improvements, we conducted paired Wilcoxon signed-rank tests over 150 test examples, comparing STReason against each baseline across all evaluation metrics.}
\begin{table}[htbp]
\centering
\caption{Wilcoxon Signed-Rank Test p-values for STReason vs. Baselines}
\label{tab:significance}
\setlength{\tabcolsep}{4pt}
\scriptsize
\begin{tabular}{lcccccc}
\toprule
\multirow{2}{*}{\textbf{Model}} 
& \multicolumn{3}{c}{\textbf{Overall Performance}} 
& \multicolumn{2}{c}{\textbf{Forecasting Accuracy}} \\
\cmidrule(lr){2-4} \cmidrule(lr){5-6}
& \textbf{Constraint} & \textbf{Factuality} & \textbf{Coherence} 
& \textbf{MAE} & \textbf{RMSE} \\
\midrule
Vicuna-7B-v1.5  & \textbf{$<$0.0001} & \textbf{$<$0.0001} & \textbf{$<$0.0001} & \textbf{$<$0.0001} & \textbf{$<$0.0001} \\
LLaMA-2-7B      & \textbf{$<$0.0001} & \textbf{$<$0.0001} & \textbf{$<$0.0001} & \textbf{0.0005}    & \textbf{0.0002}    \\
GPT-4           & \textbf{$<$0.0001} & \textbf{$<$0.0001} & 0.1573             & \textbf{$<$0.0001} & \textbf{$<$0.0001} \\
GPT-3.5 Turbo   & \textbf{$<$0.0001} & \textbf{$<$0.0001} & —                  & \textbf{0.0061}    & \textbf{0.0028}    \\
GPT-4o Mini     & 0.1573             & \textbf{$<$0.0001} & —                  & 0.6652             & 0.3895             \\
DeepSeek-V3     & \textbf{0.0455}    & \textbf{$<$0.0001} & —                  & 0.2268             & 0.1478             \\
\bottomrule
\end{tabular}
\end{table}

\textbf{Overall Performance Significance:} The p-values reported in Table~\ref{tab:significance} indicate that STReason significantly outperforms all baselines in constraint adherence ($p<0.05$), except GPT-4o Mini, which achieved comparable constraint scores. In terms of factuality, STReason outperformed all models with strong significance ($p<0.0001$), confirming its reliability in producing accurate, data-grounded responses. For coherence, STReason achieved perfect scores along with GPT-3.5 Turbo, GPT-4o Mini, and DeepSeek, leading to tied outcomes and non-applicable tests. However, its advantage over LLaMA, and Vicuna are statistically significant. These results collectively validate the robustness of STReason across interpretability, factual soundness, and analytical precision.

\textcolor{black}{\textbf{Forecasting Accuracy Significance:} Based on the p-values for forecasting accuracy metrics in Table~\ref{tab:significance}, STReason demonstrated statistically significant improvements in MAE and RMSE over GPT-3.5 Turbo, GPT-4, LLaMA2, and Vicuna. The differences were not statistically significant compared to GPT-4o Mini and DeepSeek, likely due to similar performance and the effect of sequence post-processing. These results reinforce STReason’s robustness in forecasting accuracy across diverse baselines.}

\subsection{Extended Appendix} \label{app:extended_appendix}
Due to space constraints, full module specifications (Appendix B.1), qualitative output comparisons (Appendix B.2), and human evaluation details (Appendix B.3) are included in the extended appendix at \url{https://anon.to/gTKy5L}.

\end{document}